\documentclass[10pt,twocolumn,letterpaper]{article}

\usepackage[pagenumbers]{cvpr} 

\usepackage[dvipsnames]{xcolor}

\usepackage{graphicx}
\usepackage{diagbox}
\usepackage[pagebackref=true,breaklinks=true,colorlinks,bookmarks=false]{hyperref}
\usepackage{amsmath}
\usepackage{amssymb}
\usepackage[labelformat=simple,subrefformat=simple]{subcaption}
\usepackage{multirow}
\usepackage{multicol}
\usepackage{diagbox}

\def\paperID{*****} 
\def\confName{CVPR}
\def\confYear{2024}

\title{Fine-grained length controllable video captioning with ordinal embeddings}

\author{Tomoya Nitta%
\thanks{This work was done when Tomoya Nitta was at Nagoya Institute of Technology.}
\\
Toshiba
\and
Takumi Fukuzawa,
Toru Tamaki\\
Nagoya Institute of Technology, Japan
}

\begin{document}
\maketitle

\begin{abstract}

This paper proposes a method for video captioning that controls the length of generated captions. Previous work on length control often had few levels for expressing length. In this study, we propose two methods of length embedding for fine-grained length control. A traditional embedding method is linear, using a one-hot vector and an embedding matrix. In this study, we propose methods that represent length in multi-hot vectors. One is bit embedding that expresses length in bit representation, and the other is ordinal embedding that uses the binary representation often used in ordinal regression. These length representations of multi-hot vectors are converted into length embedding by a nonlinear MLP. This method allows for not only the length control of caption sentences but also the control of the time when reading the caption. Experiments using ActivityNet Captions and Spoken Moments in Time show that the proposed method effectively controls the length of the generated captions. Analysis of the embedding vectors with ICA shows that length and semantics were learned separately, demonstrating the effectiveness of the proposed embedding methods.
\end{abstract}

\section{Introduction}
\label{sec:introduction}

Video captioning \cite{aafaq_video_2020,li_visual_2019} is a task that describes the content of a given video. Various methods have been proposed, and many of these are extensions of image captioning by handling temporal information about the content of the video, and related tasks are diverse, such as; dense video captioning \cite{Krishna_2017_ICCV} for detecting multiple temporal events with descriptions, video paragraph captioning \cite{song_towards_2021} for generating a paragraphs composed of multiple sentences, or video storytelling \cite{li_video_2020} producing textual summaries for events in the video.

Captioning tasks pose a challenge; it is difficult to predict the content of captions before they are generated, making them difficult for users to control. To address this issue, some approaches attempt to generate captions that align with the user's intentions \cite{chen_say_2020,zheng_intention_2019} or control signals \cite{cornia_show_2019}. 
In the field of controllable text generation and summarization,
various approaches aim to manage style, domain, and attributes in text generation \cite{ficler_controlling_2017}, summarization \cite{zhang_macsum_2023}, translation \cite{kobus_domain_2017}, and paraphrasing \cite{xue_sect_2022}. These techniques are also being applied to captioning in order to incorporate specific styles, domains, and attributes \cite{yang_visual_2023,chen_say_2020,zheng_intention_2019,deshpande_fast_2019}.

Among the style and attribute for captions of videos, this paper focuses particularly on length control. By controlling the length of the generated sentence, we can produce a suitable sentence or headline \cite{hitomi_large-scale_2019} that fits to a given space for the description. This concept of length control has been studied not only in image captioning \cite{deng2020length,ding_image_2024,zeng_conzic_2023,kastner_imageability-_2021,luo_controlling_2020,hirsch_clid_2024} but also in fields such as length controllable generation \cite{ficler_controlling_2017,hitomi_large-scale_2019,chai_fast_2022}, text summarization \cite{kikuchi_controlling_2016,zhang_macsum_2023,bian_controllable_2019,makino_global_2019,takase_positional_2019,jie_prompt-based_2023,fan_controllable_2018,liu_controlling_2018,saito_length-controllable_2020,yu_lenatten_2021,he_ctrlsum_2022,liu_length_2022}, translation \cite{takeno_controlling_2017} and paraphrasing \cite{xue_sect_2022}. However, the main approach in previous research is coarse control of a few length levels, and there have been hardly any attempts to fine-grained control on the length of generated captions. Also, it is difficult to use off-the-shelf text decoders pre-trained on a large dataset because many methods propose special dedicated architectures. Given these conditions, our work focuses on a method for more precise control of length, which is efficient and straightforward.

Various methods have been proposed for video captioning \cite{chen_deep_2019,islam_exploring_2021,li_visual_2019,qasim_dense_2023,abdar_review_2023,vaishnavi_video_2024,yousif_exploring_2023,jain_video_2022}, but research on length control in video captioning has not yet been attempted.
If it becomes possible to control so that the temporal length (duration) of the generated description (specifically, the time it takes for a text-to-speech system to read the sentence) matches the length of the video, it would have a significant impact on the generation of video narration and storytelling \cite{habibian_videostory_2014,huang_visual_2016,gella_dataset_2018,li_informative_2019,li_video_2020,hu_what_2020,su_bert-hlstms_2021}.
Currently, there are no methods for length-controllable text generation or captioning with specific duration constraints. Thus, the goal of this paper is to introduce an effective approach that allows for the control of caption length in the same manner, covering both textual length (measured by token count) and temporal length (measured by duration in seconds).

The proposed method employs an autoregressive (AR) text decoder with length embedding. Although there are various methods to control text length in AR decoding, we extend the length embedding \cite{deng2020length,ding_image_2024} for a more fine-grained length control.
This method can be used easily and effectively to incorporate length information into existing AR decoders. Moreover, it allows for the use of an existing video model as a visual encoder for an encoder-decoder captioning model. Consequently, length controllability can be introduced with minimal architecture changes to the encoder and decoder.

In the experiment, we used a greedy search for decoding instead of a common beam search or sampling, to evaluate the effect of length control itself.
By specifying various lengths, we generated video captions to analyze the behavior of the proposed method and assess the impact of length on generated captions. In addition, we used Independent Component Analysis (ICA) \cite{yamagiwa2023discovering} to analyze the length embedding obtained. This provides important clues for understanding how various independent components express different lengths and vocabularies. Furthermore, we will explain a training procedure and a caption generation experiment that specifies the duration of synthesized speech of the generated sentence by using a text-to-speech application, which allows us to control the temporal length of the caption.

\section{Related work}
\label{sec:related work}

Controllable length generation of text has been studied in the last decade
for text summarization \cite{kikuchi_controlling_2016,fan_controllable_2018,he_ctrlsum_2022},
translation \cite{takeno_controlling_2017},
and paraphrasing \cite{xue_sect_2022},
as well as
image captioning \cite{deng2020length,ding_image_2024,zeng_conzic_2023,kastner_imageability-_2021,luo_controlling_2020,hirsch_clid_2024}.

\paragraph{Autoregressive vs non-autoregressive decoders.}
Decoders often use an autoregressive (AR) model, but it has been pointed out \cite{deng2020length,ding_image_2024} that decoding is sequential and thus the computational complexity increases linearly. Therefore, some works \cite{zeng_conzic_2023,deng2020length,ding_image_2024} used a non-AR decoder such as BERT \cite{Devlin-ACL2019-BERT}, and it is suitable for length control because we simply prepare an input sentence with a specified length to obtain an output sentence of the same length. However, there are two issues. The first is that the generated captions include PAD tokens and hence may be shorter than the specified length, and a specific process to prevent this is necessary. The second is that approaches with non-AR decoder iterate a sentence decoding process. 
Starting with a sentence filled with MASK tokens of a specified length,
LaBERT \cite{deng2020length} and LaNAR-BERT \cite{ding_image_2024} iteratively refine the generated sentence \cite{ghazvininejad_mask-predict_2019,lee_deterministic_2018}, while Gibbs-BERT \cite{zeng_conzic_2023} predicts words iteratively using Gibbs sampling. As a result, the non-sequential characteristic of non-AR decoders is compromised.
Another drawback of the non-AR approach for this work is the inability to control the reading duration of the generated caption, as the duration does not match the word count.

\paragraph{Controlling the probability of EOS.}
There are several approaches to generate a sentence of a specified length with an AR decoder. One of them is to increase the probability of the EOS token as the length of the generated sentence approaches the specified length \cite{deng2020length,ding_image_2024,liu_length_2022}. This guarantees that the EOS token is produced reliably when the generated sentence reaches the specified length. Consequently, the generation of a sentence exceeding the specified length is not an issue; however, the generation of a sentence that falls short of the specified length remains problematic. Another problem is that it is not easy in practice \cite{deng2020length,ding_image_2024} to incorporate this procedure into a common decoding process, such as a beam search or sampling.

\paragraph{Prefix learning.}
Another approach for length control decoding is to use prefix learning to prepend length control prompts to the transformer decoder \cite{luo_controlling_2020,takeno_controlling_2017,fan_controllable_2018,he_ctrlsum_2022,chai_fast_2022,zhang_macsum_2023}. A similar idea has been considered for controlling or learning the initial conditions for LSTM \cite{liu_controlling_2018,takeno_controlling_2017,bian_controllable_2019,kikuchi_controlling_2016}. 
However, the control information is only introduced at the beginning, whose importance may gradually decrease as the sentence is generated.

\paragraph{Length embedding.}
Yet another approach is to add a length embedding vector to each word embedding that is input to the transformer decoder. This is different from prefix learning because a length embedding is added at each word that is sequentially generated, so there is the advantage of being able to continue to give length control information to the decoder until the whole sentence is generated.
There are two ways of embedding lengths: one is to add an embedding of the remaining length $T-t$ up to the specified length $T$ for the $t$-th generated word \cite{luo_controlling_2020,kikuchi_controlling_2016,yu_lenatten_2021}, and the other is to add a same length embedding to each word embedding regardless of the remaining length \cite{ding_image_2024,deng2020length}. The former would be effective, since it can give the decoder at each generation step the information on how many words remain to be generated. However, there is the disadvantage that it is necessary to modify the decoding process, which prevents the use of off-the-shelf decoder models. Also, it is neither easy nor cheap to compute the remaining temporal length to control the duration to read the generated sentence.
Therefore, we adopt the constant length embedding \cite{ding_image_2024,deng2020length}.

\paragraph{Length levels.}
Most of the previous work on length control managed lengths in units of rough length levels. Some studies represent a certain range of lengths as one level (e.g., 10 to 20 words as a single length-level) and control length with 4 or 5 discrete levels \cite{he_ctrlsum_2022,ficler_controlling_2017,ding_image_2024,deng2020length}, 3 levels such as long, medium, and short \cite{zhang_macsum_2023,xue_sect_2022}, or even 2 levels such as long and short \cite{chai_fast_2022}, and few exceptions have attempted a fine-grained length control 
(a non-AR method \cite{zeng_conzic_2023} with $n=5,8,12,15$, or LSTM \cite{luo_controlling_2020} with $n=7, \ldots, 28$). 
Moreover, all of these methods use one-hot vectors for representing length with a linear embedding matrix, similar to word embeddings, making it difficult to achieve detailed length control. Instead, we propose two nonlinear embeddings; ordinal embedding using multi-hot vectors used in ordinal regression, and bit embedding using vectors that represent length in binary representation.

\paragraph{Use of LLM.}
Previous work on length controllable generation or captioning, including this paper,
mainly considers an encoder-decoder architecture. However, due to recent progress in LLM \cite{chang_survey_2024, kalyan_survey_2024, hadi_large_2023, zhao_survey_2023, yang_harnessing_2024}, captions of a particular length can be generated using an LLM. Caption Anything \cite{wang_caption_2023} first generates captions with a captioning model and then revises them with an existing LLM (chatGPT \cite{yang_harnessing_2024}) by specifying the length or style using appropriate prompts. Jie et al. \cite{jie_prompt-based_2023} designed a reward to generate text of the length specified in the prompt, and learned LLM using reinforcement learning.
Although such methods are taking advantage of the powerful capabilities of LLM, 
a combination of video paragraph captioning and length controllable text summarization could also be a possible solution like in \cite{song_towards_2021}.
However, in this study, we develop a dedicated method because we control not only the word count of a generated sentence, but also the duration to read the sentence, which is not taken into account by LLM's word generation (currently, at least).
In addition, the performance of the captioning model that generates the caption to be refined may affect the subsequent refinement by LLM.

\section{Length embedding}

This section outlines video captioning
and length embedding methods.

A video encoder takes the video input $\boldsymbol{V}$ and produces a feature vector $\boldsymbol{h}$ as output;
\begin{equation}
    \boldsymbol{h} = \mathrm{Encoder}(\boldsymbol{V}).
\end{equation}
An autoregressive text decoder generates $i$-th token $\boldsymbol{y}_{i} \in \{0,1\}^v$,
a one-hot vector with vocabulary size $v$,
by combining the video feature $\boldsymbol{h}$
with an embedding vector $\boldsymbol{x}_{i-1} \in \mathbb{R}^d$ of $(i-1)$-th token;
\begin{equation}
    \boldsymbol{y}_{i} = \mathrm{Decoder}(\boldsymbol{h}, \boldsymbol{x}_{i-1}).
\end{equation}
The embedding layer takes as input $i$-th token $\boldsymbol{y}_{i}$
and length $k = \{1,\ldots,K\}$, and outputs an embedding;
\begin{align}
    \boldsymbol{x}_{i} 
    &= 
    \boldsymbol{W}^{\top}_{w} \boldsymbol{y}_{i} 
    + \boldsymbol{e}_{l, k} 
    + \boldsymbol{e}_{p, i},
    \label{eq:length}
\end{align}
where
$\boldsymbol{W}_{w} \in \mathbb{R}^{v \times d}$ is a word embedding matrix,
$\boldsymbol{e}_{p,i} \in \mathbb{R}^d$ is the positional embedding for $i$-th token,
and 
$\boldsymbol{e}_{l,k} \in \mathbb{R}^d$ is the length embedding vector of length $k$.
As in \cite{deng2020length,ding_image_2024},
the length of a ground-truth caption is used as $k$ for each training sample during training, 
and the specified length is given as $k$ to generate each word for inference.

The following subsections discuss how the length embedding is implemented.

\subsection{Length Level Embedding}
\label{sec:length_level_embed}

Length-level embedding \cite{deng2020length,ding_image_2024}
partitions length into $K$ levels and assigns
a length level $k$ of interval $[L_{k}, L_{k+1}-1]$ that satisfies $L_{k} \le L < L_{k+1}$
to a text $S=\{s_i\}^{L}_{i=1}$ with length of $L$.
The length level embedding $\boldsymbol{e}_{l, k}$ 
is computed with a trainable embedding matrix $\boldsymbol{W}_l \in \mathbb{R}^{K \times d}$
for length level $k$;
\begin{equation}
    \boldsymbol{e}_{l, k} = \boldsymbol{W}^{\top}_l \boldsymbol{t}_{l, k}
     \in \mathbb{R}^d,
    \label{eq:length_level}
\end{equation}
where $\boldsymbol{t}_{l, k}$ is a one-hot vector representing the length level $k$,
\begin{equation}
  \boldsymbol{t}_{l, k} = [0, \ldots, 0, \overset{\substack{k\\ \vee}}{1}, 0, \ldots, 0]
  \in \mathbb{R}^{K}.
\end{equation}
In our experiments, we assign a single integer to each interval, which means that the number of intervals is the same as the maximum length. This allows a fine-grained length control and is different from the previous work \cite{deng2020length,ding_image_2024} that used few length levels.
\footnote{
An example of $K=4$ levels \cite{deng2020length} is
[1, 9], [10, 14], [15, 19], [20, 26].
}

This approach simply extends the idea of word embedding to an integer length level $k$.
It looks reasonable because, like words, each length would have a unique embedding. However, this method is not necessarily suitable for representing ordinal numbers like lengths. Our experimental results show that the length-level embedding does not necessarily reflect the length difference in the embedding similarity (e.g., 
$\boldsymbol{e}_{l,5}$ and $\boldsymbol{e}_{l,6}$ are not necessarily more similar to each other than $\boldsymbol{e}_{l,5}$ and $\boldsymbol{e}_{l,8}$). This issue can be particularly problematic in tasks where fine-grain controlling length is essential. Hence,  we propose more appropriate methods for length embedding
in the following subsections.

\subsection{Bit Embedding}
\label{sec:bit_embed}

To ensure that the embeddings reflect an order relation, it is essential that the columns of the embedding matrix interact with each other. In this paper, we propose two methods of representing the length as a vector not in one-hot form as above, but in multi-hot form.

First, we propose a method called \emph{bit embedding}, that converts the binary representation of the length into a multi-hot vector as follows;
\begin{equation}
  \boldsymbol{t}_{bl, k} = [t_{\log_2 K}, \ldots, t_2, t_1]
  \in \mathbb{R}^{\log_2 K},
\end{equation}
where $t_i$ represents the $i$-th bit in the binary representation of the decimal number $k$,
and $K$ is the maximum length.

Furthermore, we introduce a nonlinear embedding with an MLP instead of using a linear embedding matrix;
\begin{equation}
  \boldsymbol{e}_{bl, k} = \mathrm{MLP} (\boldsymbol{t}_{bl, k})
   \in \mathbb{R}^d.
  \label{eq:bit}
\end{equation}
The reason for using nonlinear embedding is that in the linear case, the sum of the columns of the embedding matrix corresponding to the bit representation does not necessarily correspond to the sum of different binary bits.
Consider $\boldsymbol{t}_{bl, 7} = [0, 1, 1, 1]$ for example, $2^0+2^1+2^2=1+2+4=7$, but there is no constraint for $\boldsymbol{e}_{l,0} + \boldsymbol{e}_{l,1} + \boldsymbol{e}_{l,2}$ to be close to $\boldsymbol{e}_{l,3}$ (of length 8 with $\boldsymbol{t}_{bl, 8} = [1, 0, 0, 0]$), and a simple linear combination alone shows the problem of not being able to guarantee a correspondence between the representation of the bit and the embedding vectors. Using a non-linear MLP, the result of the first layer (embedding matrix) may be further transformed to satisfy the above constraints. However, in experiments, the similarity between embedding vectors tends to be greatly influenced by the higher bits. Therefore, we will show a method to avoid this problem next.

\subsection{Ordinal Embedding}
\label{sec:ordinal_embed}

Here we propose the \emph{ordinal embedding};
it uses a binary representation of ordinal regression \cite{tutz_ordinal_2022}
in which an ordinal response is managed by the binary responses
in the following form;
\begin{equation}
    \boldsymbol{t}_{ol, k}
    = [\underbrace{1, \ldots, 1}_{k}, \underbrace{0, \ldots, 0}_{K-k}]
    \in \mathbb{R}^{K},
\label{eq:OE}
\end{equation}
that is, first $k$ elements are 1 and the rest $K-k$ are zero.
This is used as input for obtaining a nonlinear embedding with MLP;
\begin{equation}
  \boldsymbol{e}_{ol, k} = \mathrm{MLP} (\boldsymbol{t}_{ol, k})
   \in \mathbb{R}^d.
  \label{eq:ordinal}
\end{equation}

In contrast to bit embedding, ordinal embedding is the sum of the columns that corresponds to the ordinal length representation. In a linear case, for example, the length representation of $k=2$ is [1,1,0,0], which corresponds to
$\boldsymbol{e}_{l,1} + \boldsymbol{e}_{l,2}$.
On the other hand, the length representation of $k=3$ is [1,1,1,0], which becomes $\boldsymbol{e}_{l,1} + \boldsymbol{e}_{l,2} + \boldsymbol{e}_{l,3}$. Thus, the order constraint is naturally introduced; the difference between the length representations $\boldsymbol{t}_{ol, k}$ increases with length, and the corresponding length embeddings $\boldsymbol{e}_{ol, k}$ differ as with the additional terms. In the proposed method, a more detailed length representation is obtained using nonlinear embedding by MLP.

\section{Experiment}
\label{sec:experiment}

\subsection{Dataset}
\label{sec:dataset}
\label{sec:spoken_mit}

\begin{figure}[t]
    \centering

    \subcaptionbox{\label{fig:activitynet_hist}}{
        \includegraphics[width=0.47\linewidth]{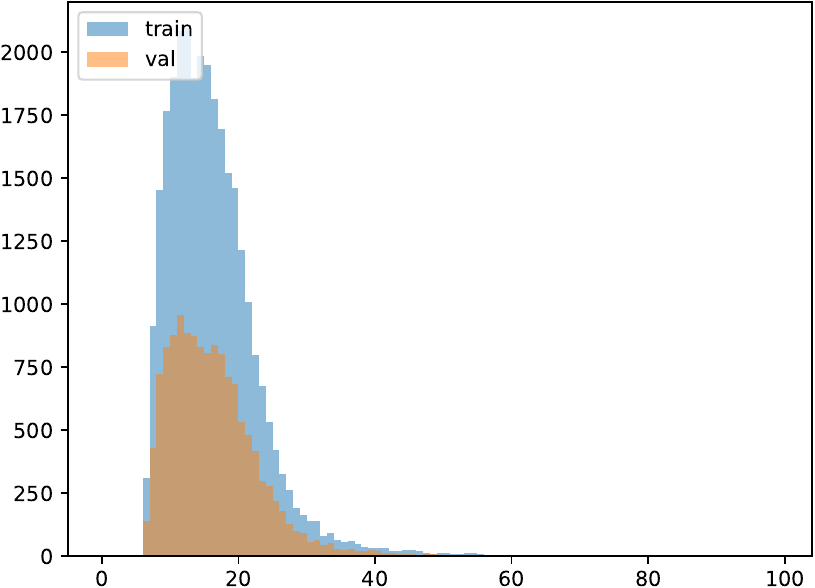}
    }
    \subcaptionbox{\label{fig:spoken_mit_hist}}{
        \includegraphics[width=0.47\linewidth]{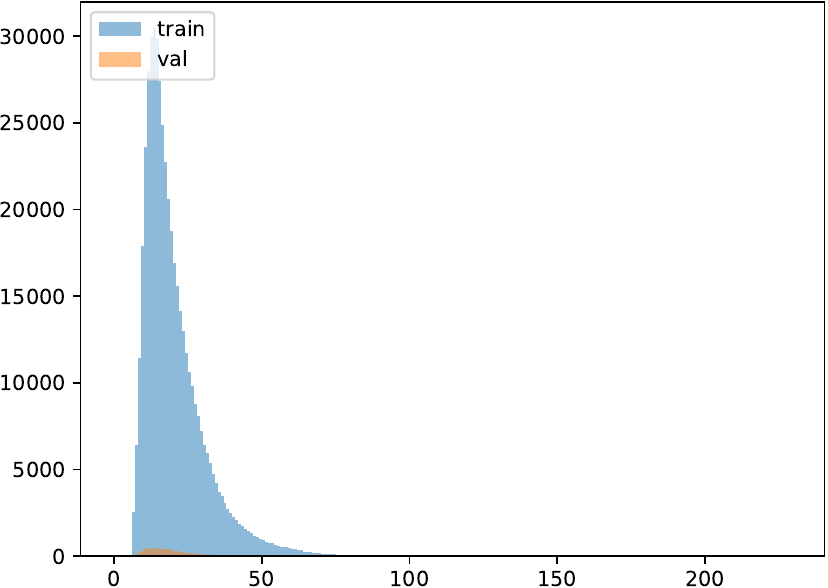}
    }

  \caption{
      Histograms of the lengths of annotated captions of
      \subref{fig:activitynet_hist} ActivityNet Captions and 
      \subref{fig:spoken_mit_hist} Spoken MiT.
  }

\end{figure}

In our experiments, we used two datasets: ActivityNet Captions \cite{Krishna_2017_ICCV} and Spoken Moments in Time (MiT) \cite{monfortmoments, 9609554, Monfort_2021_CVPR}. As mentioned in \cite{hirsch2022clid}, a training dataset must contain a sufficient number of captions with the desired length for effective length embedding. The length statistics of these datasets, using the GPT2 tokenizer \cite{Radford_2019_GPT2}, are shown in Table \ref{tab:dataset}. Note that hereafter we define ``length'' as the ``number of tokens'' (or token counts) to simplify experiments and analysis because subword tokenizers have been commonly used.

\begin{table*}[t]
    \centering

    \caption{Dataset statistics on length (token counts) of captions. 
    Each column shows the minimum, maximum, mean and std of the number of tokens in a single sentence.
    Note that numbers may differ from those presented in 
    \cite{Krishna_2017_ICCV, monfortmoments, 9609554, Monfort_2021_CVPR}
    because some videos have become unavailable since their release.
    }
    \label{tab:dataset}

    \begin{tabular}{cc|cccc}
        dataset &  split & min & max & mean $\pm$ std  & \#sentences \\ \hline
        \multirow{2}{*}{ActivityNet Captions}
        & train & 6 & 84 & 15.80 $\pm$ 6.89 & 29520 \\
        & val & 6 & 100 & 15.94 $\pm$ 7.08 & 13632 \\ \hline
        \multirow{2}{*}{Spoken Moments in Time} 
        & train & 2 & 224 & 19.80 $\pm$ 10.87 & 479632 \\
        & val & 4 & 128 & 19.40 $\pm$ 10.37 & 8072 \\ \hline
    \end{tabular}

\end{table*}

ActivityNet Captions is a dataset that includes YouTube videos of a few minutes, annotated with event intervals (start and end timestamps) and corresponding caption texts.
Figure \ref{fig:activitynet_hist} shows the histogram of the lengths of the caption text. As observed in Table \ref{tab:dataset}, the length distributions between the train and validation datasets are not significantly different. The minimum length for both is 6, therefore, controlling lengths of 5 or fewer would require the model to handle lengths not seen during training. In addition, it is likely that a model trained on this dataset will generate captions with average length. 
To manage lengths that are considerably shorter than the average, the model needs to learn length embeddings from data with limited length ranges.

Spoken Moments in Time (Spoken MiT) is a dataset comprising short 3-second videos, sourced from various social media platforms, with annotated descriptions. Some original description texts lack a period (``.'') at the end of the sentence, hence we added a period to those sentences to align the experimental setting with ActivityNet Captions, which have periods at the end of all caption texts. Figure \ref{fig:spoken_mit_hist} shows a histogram of the lengths of the description texts. 
This indicates that this dataset is more suitable for learning length embeddings than ActivityNet Captions, since this dataset exhibits a broader range of lengths.
The minimum length is 2 tokens, which suggests that some original texts consist of only one token
because adding a period increases the token count by one.
Therefore, the dataset does not suffer from the insufficient data issue to obtain length embeddings for shorter lengths.

\subsection{Experimental Settings}
\label{sec:Experimental Settings}

In our experiments, we used a baseline video captioning model, which uses VideoMAE \cite{tong2022videomae} as the video encoder and GPT2 \cite{Radford_2019_GPT2} as the text decoder. Both the encoder and decoder are pretrained;
VideoMAE uses ViT-B \cite{Dosovitskiy_ICLR2021_ViT_Vision_transformer} as a backbone, pretrained on the Kinetics-400 action recognition dataset \cite{kay_arXiv2017_kinetics400},
and the used GPT2 is the smallest model with 124M parameters, pretrained on the WebText \cite{Radford_2019_GPT2}.
The CLS token from the video encoder's output is used as input feature $\boldsymbol{h}$
for the text decoder's cross-attention.

The length of a caption is determined by the number of tokens obtained by the GPT2 tokenizer. This token count includes all word tokens, including periods (``.'') but excluding special ones such as BOS, EOS, UNK, and PAD. The maximum length is set to 256. Thus, the dimensionality of the length representation vectors $\boldsymbol{t}$ is 256 for the length level and the ordinal embedding, and 8 for the bit embedding. The bit and ordinal embeddings $\boldsymbol{e}_{l}$ were obtained using a 3-layer MLP, with hidden dimensions
of (64, 256) for the bit embedding and (512, 512) for ordinal embedding, based on preliminary experiments.

\subsubsection{Training}
As is typical in action recognition tasks,
the video encoder takes an input clip consisting of 16 frames, which were uniformly sampled from an annotated event interval for ActivityNet Captions videos and the entire video for Spoken MiT. Clips shorter than 16 frames were excluded from the training. The short edge of the clip was randomly scaled to fall within the range [256, 320], preserving the original aspect ratio. The clip was then randomly cropped to a size of $224 \times 224$, followed by a random horizontal flip. AdamW \cite{loshchilov2018decoupled} optimizer was used, with a learning rate of 1e-4 and a weight decay of 5e-4. The batch size is set to 4 per GPU, with parallel processing performed using PyTorch's distributed data parallel on 4 GPUs. Gradients were updated every 8 batches.

\subsubsection{Inference and generation}
\label{sec:inference}

The input clip was prepared in the same way as for the training. The short edge of the clip was scaled to 256 while maintaining the original aspect ratio, followed by a central crop of size $224 \times 224$.
For caption generation, we employ a greedy decoding approach, instead of a common beam search, to evaluate the effect of length control itself, 
although the quality metrics (see below) may not match those from previous work.
The specified lengths (called \emph{target length} in this section) for the arbitrary length test \cite{liu_length_2022} were set to 5 and 20, and the length of the ground truth caption (denoted as GT length in the following) was used for the gold length test \cite{liu_length_2022,saito_length-controllable_2020}.
The upper limit for the length of captions generated by greedy decoding was set to the target length plus 10 tokens. Therefore, for target lengths of 5 and 20, the longest possible captions are 15 and 30, respectively, and the generation process stops at this upper limit.
We used BLEU-4 \cite{papineni-etal-2002-bleu}, ROUGE-L \cite{lin-2004-rouge}, METEOR \cite{banerjee-lavie-2005-meteor}, and CIDEr \cite{Vedantam_2015_CVPR} as metrics for evaluation of captions.
However, the mean and standard deviation of the lengths were used as the main metrics to evaluate the lengths of the generated captions.
Note that the length (or token counts) of generated captions is calculated using the GPT2 tokenizer. This is done by decoding the generated token sequence to obtain a plain text, and then re-tokenizing it with the same tokenizer to count tokens. This helps to exclude irrelevant tokens, such as PAD tokens, from the generated token sequence.

\subsection{Result}
\label{sec:result}

\begin{table*}[t]
    \centering
    
    \caption{Results on ActivityNet Captions for different target lengths for three types of embedding.}
    \label{tab:activitynet}

    \begin{tabular}{c|c|cccccc}
      embedding  & \parbox{3em}{\centering target length} & CIDEr & METEOR & BLEU-4 & ROUGE-L & tokens & difference $\downarrow$\\ \hline
      \multirow{3}{*}{
        \parbox{4em}{\centering Length Level}
      }
      & 5   & 0.1466 & 0.0764 & 0.0185 & 0.1880 & 14.99 $\pm$ 0.1357 & 9.99 \\
      & 20  & 0.2101 & 0.1077 & 0.0352 & 0.2058 & 22.60 $\pm$ 2.216  & 2.60 \\
      & GT  & 0.4028 & 0.1055 & 0.0449 & 0.2243 & 18.18 $\pm$ 9.792  & \\ \hline
      \multirow{3}{*}{Bit}  
      & 5   & 0.2413 & 0.0710 & 0.0185 & 0.1886 & 8.819 $\pm$ 0.8445 & 3.819 \\
      & 20  & 0.2411 & 0.1075 & 0.0365 & 0.2084 & 19.95 $\pm$ 1.102  & 0.05 \\
      & GT  & 0.4282 & 0.1062 & 0.0464 & 0.2247 & 15.83 $\pm$ 7.126  & \\ \hline
      \multirow{3}{*}{Ordinal}   
      & 5   & 0.2046 & 0.0608 & 0.0117 & 0.1710 & 7.048 $\pm$ 0.8459 & 2.048 \\
      & 20  & 0.2421 & 0.1068 & 0.0360 & 0.2089 & 19.71 $\pm$ 1.338  & 0.29\\
      & GT  & 0.4255 & 0.1049 & 0.0464 & 0.2241 & 15.68 $\pm$ 7.275  & \\ \hline
    \end{tabular}
\end{table*}

\begin{table*}[t]
    \centering

    \caption{Results on Spoken MiT for different target lengths for three types of embedding.}
    \label{tab:spoken_mit}

    \begin{tabular}{c|c|cccccc}
      embedding  & \parbox{3em}{\centering target length} & CIDEr & METEOR & BLEU-4 & ROUGE-L & tokens & difference $\downarrow$ \\ \hline
      \multirow{3}{*}{
        \parbox{4em}{\centering Length Level}
      }
      & 5   & 0.1307 & 0.0499 & 0.0027 & 0.1665 & 4.718 $\pm$ 1.010  & 0.282 \\
      & 20  & 0.3245 & 0.1227 & 0.0536 & 0.2534 & 20.03 $\pm$ 0.6008 & 0.03\\
      & GT  & 0.5948 & 0.1284 & 0.0581 & 0.2715 & 20.26 $\pm$ 11.75  & \\ \hline
      \multirow{3}{*}{Bit}
      & 5   & 0.2869 & 0.0773 & 0.0159 & 0.2195 & 8.016 $\pm$ 0.1380 & 3.016 \\
      & 20  & 0.3306 & 0.1222 & 0.0526 & 0.2514 & 19.97 $\pm$ 0.2000 & 0.03\\
      & GT  & 0.6082 & 0.1294 & 0.0585 & 0.2693 & 19.87 $\pm$ 10.62  & \\ \hline
      \multirow{3}{*}{Ordinal}
      & 5   & 0.1491 & 0.0518 & 0.0025 & 0.1614 & 5.155 $\pm$ 0.3624 & 0.155 \\
      & 20  & 0.3254 & 0.1218 & 0.0523 & 0.2506 & 20.12 $\pm$ 0.3417 & 0.12 \\
      & GT  & 0.5984 & 0.1293 & 0.0578 & 0.2691 & 19.74 $\pm$ 10.52 \\ \hline
    \end{tabular}
\end{table*}

\begin{table*}[t]
    \centering
    
    \caption{Results on ActivityNet Captions for ordinal embedding with ViViT \cite{Arnab_2021_ICCV_ViVit} as the encoder.}
    \label{tab:activitynet_vivit}

    \begin{tabular}{c|c|cccccc}
      embedding  & \parbox{3em}{\centering target length} & CIDEr & METEOR & BLEU-4 & ROUGE-L & tokens & difference $\downarrow$\\ \hline
      \multirow{3}{*}{
        \parbox{4em}{\centering Length Level}
      }
      & 5   & 0.0544 & 0.0549 & 0.0081 & 0.1315 & 14.9997 $\pm$ 0.0428 & 9.9997 \\
      & 20  & 0.1469 & 0.1064 & 0.0300 & 0.1906 & 25.4914 $\pm$ 2.6393 & 5.4914 \\
      & GT  & 0.3245 & 0.1044 & 0.0377 & 0.2082 & 20.9577 $\pm$ 9.8243 & \\ \hline
      \multirow{3}{*}{Bit}  
      & 5   & 0.2143 & 0.0707 & 0.0158 & 0.1786 & 9.8684 $\pm$ 1.1372 & 4.8684 \\
      & 20  & 0.1967 & 0.1062 & 0.0320 & 0.1959 & 21.7186 $\pm$ 4.5105 & 1.7186 \\
      & GT  & 0.3715 & 0.1039 & 0.0421 & 0.2112 & 16.7950 $\pm$ 9.8225  & \\ \hline

      \multirow{3}{*}{Ordinal}   
      & 5   & 0.1867 & 0.0585 & 0.0096 & 0.1612 & 7.356 $\pm$ 0.7451 & 2.356 \\
      & 20  & 0.2225 & 0.1074 & 0.0370 & 0.1978 & 20.84 $\pm$ 1,1143  & 0.84\\
      & GT  & 0.3992 & 0.1056 & 0.0480 & 0.2125 & 16.245 $\pm$ 6.8931  & \\ \hline
    \end{tabular}
\end{table*}

\subsubsection{Quantitative evaluation on length}
\label{sec:Quantitative evaluation}

Table \ref{tab:activitynet} shows the results of ActivityNet Captions. The length level embedding failed to control for a target length of 5, with all generated sentences reaching maximum length (5 + 10 = 15). This is due to the lack of short captions in the training data, which results in ineffective length control. For target lengths of 20 and GT length, the control improves but is still off by 2 tokens.
Bit and ordinal embeddings demonstrate better control. For a target length of 5, the average token lengths vary by only 2 or 3 tokens from the target length. Representing lengths with multi-hot vectors allows the embedding of unseen lengths by leveraging their relationship with the lengths present in the training data. 
For a target length of 20, the performance of both embeddings is comparable and the difference to the target length is less than one.

Table \ref{tab:spoken_mit} shows the results of Spoken MiT.
For the target length of 20, all embeddings produce captions that are approximately the target length.
For the target length of 5, unlike the results from ActivityNet Captions, the length-level embedding generates captions that are close to the target length, demonstrating its capability to generate short captions only when provided with enough training data.
Bit embedding, despite having sufficient training data, struggles to control length, and generated captions with lengths exceeding the target by about 3 tokens.
Ordinal embedding shows effective length control in both datasets, indicating that the embedding can adapt to unseen length through correlation learning during training.

In summary, bit and ordinal embeddings outperform length-level embedding in length control, including unseen short lengths, but ordinal embedding is the most effective overall.
Note that this finding may hold for another model variant. As an example, Table \ref{tab:activitynet_vivit} shows results with a model that uses ViViT \cite{Arnab_2021_ICCV_ViVit} as the encoder instead of VideoMAE. We observe a similar trend in the length of the captions generated with each embedding, and the ordinal embedding performs the best in this case as well.

\subsubsection{Quantitative evaluation on metrics}

When comparing the evaluation metrics of the results of ActivityNet Captions in Table \ref{tab:activitynet}, there are no significant differences between METEOR, BLEU-4, and ROUGE-L for GT length. However, CIDEr slightly improves by about 0.02 points with both bit and ordinal embeddings compared to length-level embedding.
A notable drop in the metric is observed for a target length of 5, probably due to the large length differences between the reference and generated captions.
METEOR, BLEU-4, and ROUGE-L calculate scores based on n-gram matching between the reference and the generated captions. Therefore, a difference in length likely leads to a decrease in the score. CIDEr calculates scores based on the cosine similarity of TF-IDF weighted n-grams, which is affected by changes in word frequency patterns between reference and generated captions.
For a target length of 20, all embeddings show a decrease in CIDEr scores. However, the decrease is less noticeable in METEOR, BLEU-4, and ROUGE-L, likely due to the smaller length difference.

Similar trends can be observed in the results of Spoken MiT, as shown in Table \ref{tab:spoken_mit}. Although the metrics fall for target length of 5, there is a significant drop in CIDEr for a target length of 20, even though there is sufficient training data.

\begin{figure*}
  \begin{subfigure}{\textwidth}
    \centering
    \includegraphics[width=0.9\linewidth]{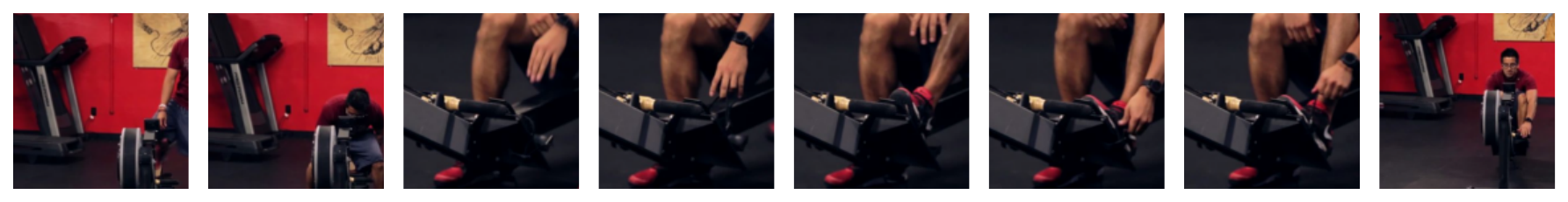}
  \end{subfigure}

  \begin{subfigure}{\textwidth}
    \centering
    \small
    \begin{tabular}{cccp{.6\linewidth}}
      \multirow{2}{*}{embedding} &
      \multicolumn{2}{c}{length} &
      \multirow{2}{*}{captions}
      \\
       & target & gen &\\ \hline
      GT (14) &  & &
      He sits back down on the machine
      and pulls back on a bar. 
      \\ \hline
      \multirow{3}{*}{
        \parbox{4em}{\centering Length Level}
      }
      & 5 & 15 &
            The man then the man is pulling
            the straps of the pedal and he's 
      \\
      & 14 & 15 &
      The man pulls the handle lever and
      adjusts the cables to the bike.......... 
      \\
      & 20 & 26 &
            The man then pulls the bike up and down and begins
            working out of the pedal.The lever. the lever. the bike... 
      \\ \hline
      \multirow{3}{*}{Bit}
      & 5 & 10 &
      The man then adjusts the straps on the machine. 
      \\
      & 14 & 13 &
           The man then begins to use the machine 
           on the machine again. 
      \\
      & 20 & 18 &
            The man then pulls the handlebars of the bike and
            begins to tighten the handlebars. 
      \\ \hline
      \multirow{3}{*}{Ordinal}
      & 5 & 6 &
      The man is working out. 
      \\
      & 14 & 14 & 
           The man then begins to work out on 
           the machine and adjusting it. 
      \\
      & 20 & 21 &
           The man then begins to work out on the machine and
           begins to pull himself back and fourth on it. 
      \\ \hline
    \end{tabular}
  \end{subfigure}

  \caption{Examples of generated captions for a video of ActivityNet Captions with different target lengths and embedding methods.}

  \label{fig:activitynet_vid_cap}
\end{figure*}

\begin{figure*}
  \begin{subfigure}{\textwidth}
    \centering
    \includegraphics[width=.9\linewidth]{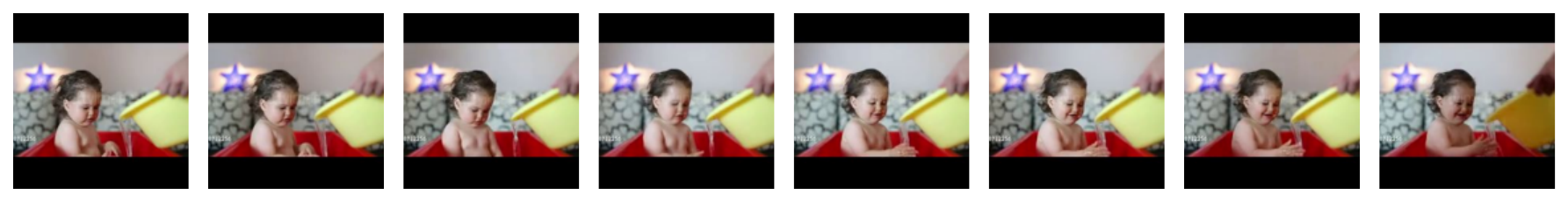}
  \end{subfigure}

  \begin{subfigure}{\textwidth}
    \centering
    \small
    \begin{tabular}{cccp{.6\linewidth}}
      \multirow{2}{*}{embedding} &
      \multicolumn{2}{c}{length} &
      \multirow{2}{*}{captions}
      \\
       & target & gen &\\ \hline
      GT (14) &  & &
           a woman helps give her daughter a bath
           using the yellow plastic tub.  
      \\ \hline
      \multirow{3}{*}{
        \parbox{4em}{\centering Length Level}
      }
      & 5 & 3 & 
      a baby. 
      \\
      & 14 & 15 &
            a baby is playing with a toy
            that is in a small bucket............ 
      \\
      & 20 & 20 &
            a baby is sitting in a bathtub with a yellow 
            sponge in front of her playing with it..... 
      \\ \hline
      \multirow{3}{*}{Bit}
      & 5 & 8 & 
      a baby is playing in a tub... 
      \\
      & 14 & 14 &
           a baby is sitting in a red
           bathtub with a yellow sponge......... 
      \\
      & 20 & 20 &
           a baby is sitting in a red bathtub the baby is
           playing with a toy in the water.... 
      \\ \hline
      \multirow{3}{*}{Ordinal}
      & 5 & 5 &
      a baby is playing.
      \\
      & 14 & 14 &
           a baby is playing in a small tub
           with a yellow rubber object. 
      \\
      & 20 & 20 &
           a baby is sitting in a red plastic tub
           with water and a pink sponge the baby is laughing. 
      \\ \hline
    \end{tabular}

  \end{subfigure}

  \caption{Examples of generated captions for a video of Spoken MiT with different target lengths and embedding methods.}
  \label{fig:spoken_mit_vid_cap}
\end{figure*}

\subsubsection{Qualitative Evaluation}

Figure \ref{fig:activitynet_vid_cap} shows an example of the captions for a video from ActivityNet Captions. Note that some captions end with a repetition of periods (``.'') generated at the end (actually, periods and special tokens alternate); however, the post-processing of the tokenizer counted them as a single token.
The length-level embedding struggle to learn the target length of 5, often reaching a maximum length of 15 (up to 10 extra length). These sentences lack a concluding period, indicating that the caption was going to continue but was truncated. For a target length of 20, the final portion of the generated caption repeats similar words. 
The bit embedding generates captions without repeating periods at the end, but the caption lengths do not align closely with the target lengths of 5.
Ordinal embedding can produce a caption of length 6 for the target length of 5. Considering that the minimum length in the ActivityNet Captions dataset is 6 (as shown in Table \ref{tab:dataset}), ordinal embedding can control for short lengths not present in the dataset.

The caption examples generated for a video from Spoken MiT are shown in Figure \ref{fig:spoken_mit_vid_cap}. 
Length-level embedding can generate a caption of length 3 for a target length of 5. This is probably because the dataset contains many short captions; however, there are still errors by one or two tokens and extra periods at the end.
Bit and ordinal embeddings generated captions that appear to describe the video content accurately with captions of a length close to the target lengths. In particular, ordinal embedding generates captions with the same length as the target length without extra periods.

\begin{figure*}[t]
    \centering

    {\small
    \begin{tabular}{ccp{.8\linewidth}}
    \multicolumn{2}{c}{length} & \multirow{2}{*}{captions} \\
      target & gen &\\ \hline
      5 & 5 & a baby is playing. \\
      6 & 6 & a baby playing in water. \\
      7 & 7 & a baby is playing with bubbles. \\
      8 & 8 & a baby is playing in a pool. \\ 
      9 & 9 & a baby is playing in a small pool. \\ 
      10 & 10 & a baby is taking a bath in a tub. \\
      11 & 11 & a baby is taking a bath in a small tub. \\ 
      12 & 12 & a baby is playing in a small tub with a toy. \\ 
      13 & 13 & a baby is playing in a small tub with a yellow sponge. \\ 
      14 & 14 & a baby is playing in a small tub with a yellow rubber object. \\ 
      15 & 15 & a baby is playing in a small tub with a yellow sponge and water. \\ 
      16 & 16 & a baby is sitting in a red plastic tub the baby is taking a bath. \\ 
      17 & 17 & a baby is sitting in a red plastic tub the baby is taking a bubble bath. \\ 
      18 & 18 & a baby is sitting in a red plastic tub the baby is taking a bath with bubbles. \\ 
      19 & 19 & a baby is sitting in a red plastic tub the baby is taking a bubble bath with bubbles. \\
      20 & 20 & a baby is sitting in a red plastic tub with water and a pink sponge the baby is laughing. \\
      \end{tabular}
    }

    \caption{Captions generated for the same video in Fig.\ref{fig:spoken_mit_vid_cap} with ordinal embedding.
    The target length ranges between 5 and 20.}
    \label{fig:5to20}

\end{figure*}

\subsubsection{Effect of target lengths on ordinal embedding}

Figure \ref{fig:5to20} shows captions generated with ordinal embedding for the same sample in Figure \ref{fig:spoken_mit_vid_cap}. The captions change as target lengths gradually increase from 5 to 20 in increments of 1.
When we vary the target tokens from 1 to 20, the generated captions match the target lengths,
while they gradually change. Specifically, at target lengths of 12, 13, and 14, the description of the same object evolves from ``toy'' to ``yellow sponge'', and then to ``yellow rubber object'', demonstrating a nuanced change in the representation of the word. Other target lengths exhibit similar trends, suggesting a systematic adjustment in the object's description.
As the token length increases (upto 20 in this figure), the generated captions start to include conjunctions like ``and'' to create compound sentences. This extends the caption length while maintaining sentence coherence.

More examples with longer target lengths and other embeddings are shown in the Appendix.

\subsubsection{Controlling for longer sentences}
\label{sec:Controlling for longer sentences}

\begin{figure*}[t]
    \centering

    \subcaptionbox{\label{fig:avg_length_anet}}{
        \includegraphics[width=0.48\linewidth]{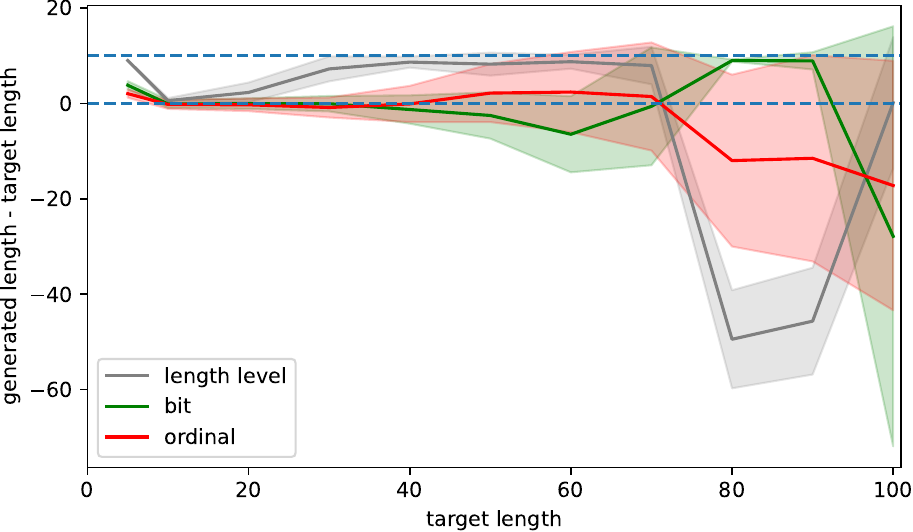}
    }
    \hfill
    \subcaptionbox{\label{fig:avg_length_smit}}{
        \includegraphics[width=0.48\linewidth]{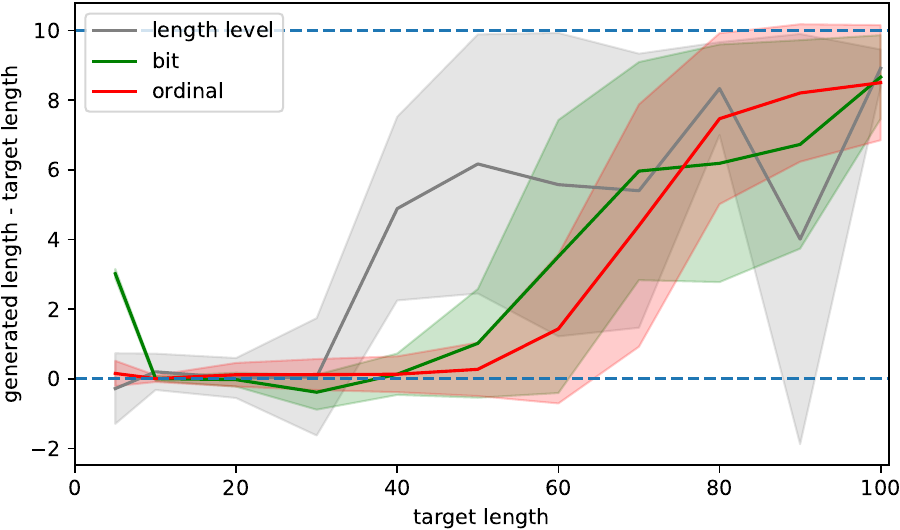}
    }

    \caption{
        The difference between the target lengths and the average lengths of generated captions
        for (a) ActivityNet Captions and (b) Spoken MiT.
    }
    \label{fig:avg_length}

\end{figure*}

Here, we show experiments on generating captions of more than 20 tokens using three types of embeddings in Figure \ref{fig:avg_length}. The solid lines show the (signed) difference between the average generated length and the target length, ranging from 5 to 100, with the standard deviations shown in the corresponding shading colors. The dashed $y=0$ line represents the case where the generated length equals the target length, while the $y=+10$ line indicates the maximum length for generation. As stated in Section \ref{sec:inference}, if a generated caption exceeds the target length by 10, it is truncated.

Figure \ref{fig:avg_length_anet} shows the results of ActivityNet Captions.
The length of captions generated by length-level embedding tends to deviate from the target length when it exceeds 20, and the length control becomes extremely unstable, especially when the target length is 80 or more. This instability is due to the fact that, as can be seen in Table \ref{tab:dataset}, the maximum length in this dataset is 84, with no data exceeding this length. Bit embedding is also unstable for target lengths of 80 or more, but in the shorter range, the standard deviation is smaller than that of length-level embedding, and generated captions tend to be slightly shorter. Ordinal embedding, while having standard deviations similar to that of bit embedding, generates captions that match the target length up to around 70. 
Moreover, despite the absence of data with a target length of 80 or more, it remains relatively stable compared to bit and length-level embeddings, 
while progressively reducing the generated length and increasing the standard deviation.

Figure \ref{fig:avg_length_smit} shows the results of Spoken MiT.
For ordinal embedding, controlling length within the range $[5, 50]$ results in captions with an average length near the target. However, beyond the range, the average length gradually exceeds the target, approaching the maximum length. This is likely because, as seen in Figure \ref{fig:spoken_mit_hist}, samples longer than 50 are less frequent in the dataset. 

This finding, which is in line with Section \ref{sec:Quantitative evaluation}, suggests that learning from limited data is feasible for short sentences, but more difficult for longer ones. The length representation of the ordinal embedding in Eq.\eqref{eq:OE} often equals 1 for lower values (leftward bits) and is less likely for higher values (rightward bits). Hence, the weights corresponding to higher values receive fewer gradient updates during training than those for lower values. This could affect the learning of length embeddings for longer sentences. A potential solution could be the use of synthetically generated longer sentences, as proposed in \cite{hirsch_clid_2024}.

Bit embedding follows a pattern similar to ordinal embedding, but exhibits a larger standard deviation for longer target lengths. The length-level embedding clearly underperforms because it generates unstable length across the range of target lengths.

\subsection{Duration control}

\begin{table*}[t]
    \centering

    \caption{
        Results on Spoken MiT for different target duration (in seconds).
    }
    \label{tab:time_emb}

    \begin{tabular}{c|c|cccccc}
       embedding & \parbox{3em}{\centering target length} & CIDEr & METEOR & BLEU-4 & ROUGE-L & duration [sec] & difference $\downarrow$ \\ \hline
      \multirow{3}{*}{Ordinal}   
      & 2.0 & 0.3259 & 0.0838 & 0.0227 & 0.2402 & 1.942 $\pm$ 0.1265 & 0.058\\
      & 5.0 & 0.2729 & 0.1224 & 0.0496 & 0.2463 & 5.170 $\pm$ 0.2379 & 0.157\\
      & GT  & 0.5566 & 0.1274 & 0.0562 & 0.2684 & 4.722 $\pm$ 2.225 \\
    \end{tabular}

\end{table*}

\begin{figure*}[t]
    \centering
    \small
    \begin{tabular}{cccp{.6\linewidth}}
      \multirow{2}{*}{embedding} & 
      \multicolumn{2}{c}{duration} & 
      \multirow{2}{*}{captions} 
      \\
       & target & gen &  \\ \hline
      GT (3.7) &  &   & 
           a woman helps give her daughter a bath
           using the yellow plastic tub. (14 tokens)
      \\ \hline
      \multirow{3}{*}{Ordinal}
      & 2.0 & 2.095 & 
      a little baby is playing in a bathtub........... (11 tokens)
      \\
      & 3.7 & 3.705 &
           a little baby is sitting in a red chair
           and is being fed with a bottle. (17 tokens)
      \\
      & 5.0 & 4.841 &
           a little baby is sitting in a red chair with a yellow
           sponge and a little yellow sponge in her mouth. (22 tokens)
      \\ \hline
    \end{tabular}

    \caption{Examples of generated captions for a video of Spoken MiT with different target duration.}
    \label{tab:time_emb_sample}

\end{figure*}

This work is also motivated by the desire to control the temporal length (i.e., duration) of the generated description, specifically the time (in seconds) it takes for a text-to-speech system to read the sentence. This section focuses on experiments for controlling the duration, not the number of tokens, using the Spoken MiT dataset and ordinal embedding.

We use eSpeak \cite{espeak}, a relatively old but lightweight and user-friendly text-to-speech software, to create audio files of the generated captions. Although recent text-to-speech (TTS) research \cite{tan_survey_2021,zhang_survey_2023} has advanced greatly, we prioritize convenience and speed over voice quality for this experiment. We used the default voice settings (gender, pitch, etc.) in eSpeak and considered the duration of the created audio file as the duration of the generated caption.
The duration in seconds is discretized in 0.1-second intervals and is represented by a 256-d multi-hot vector for the ordinal embedding. This results in a duration control within the range of 0.0 to 25.6 seconds. During training, captions exceeding 25.6 seconds in duration were treated as 25.6 seconds. During generation, the maximum number of tokens in the generated captions was set to 50 for truncation.
In our evaluation, we compared target durations of 2.0 and 5.0 seconds, as well as the ground truth (GT) duration, and used the average duration, instead of token counts.

The results are shown in Table \ref{tab:time_emb} and the generated captions in Table \ref{tab:time_emb_sample}.
Compared to experiments to control token counts,
the CIDEr score for GT duration is slightly lower, while other metrics are similar. For a target duration of 2.0 and 5.0 seconds, the duration of the generated caption was close to the target duration. The significant decrease is observed in CIDEr for 5.0 seconds and in BLEU for 2.0 seconds.
This results suggests that the duration control with ordinal embedding appears effective,
similar to token length control.
The generated captions align with the target duration, and the standard deviation is relatively small compared to that for the target duration.

\section{Analysis}

\subsection{Word frequency}

As seen in Tables \ref{tab:activitynet} and \ref{tab:spoken_mit}, a significant decrease in CIDEr was observed for a target length of 20 in both datasets compared to the GT length. This decrease may be attributed to the deviation of the token distribution in the sentences from the ground-truth annotations. To illustrate this, we analyze the histograms of words in the generated captions with ordinal embedding using SpaCy \cite{spacy}.

\begin{figure*}[t]
    \centering

    \subcaptionbox{GT length \label{fig:gt_freq}}{
        \includegraphics[width=0.3\linewidth]{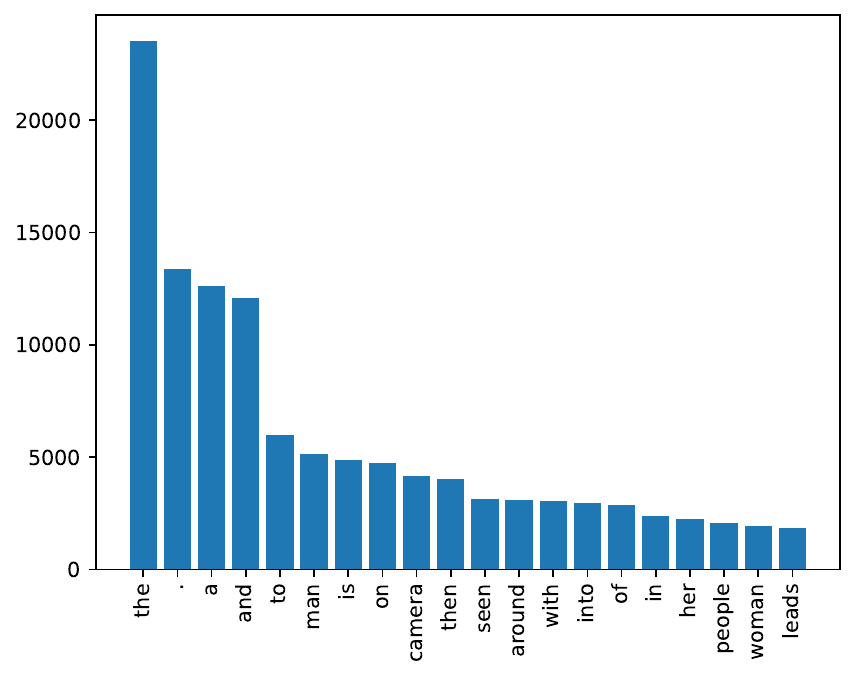}
    }
    \hfill
    \subcaptionbox{5 tokens \label{fig:min_freq}}{
        \includegraphics[width=0.3\linewidth]{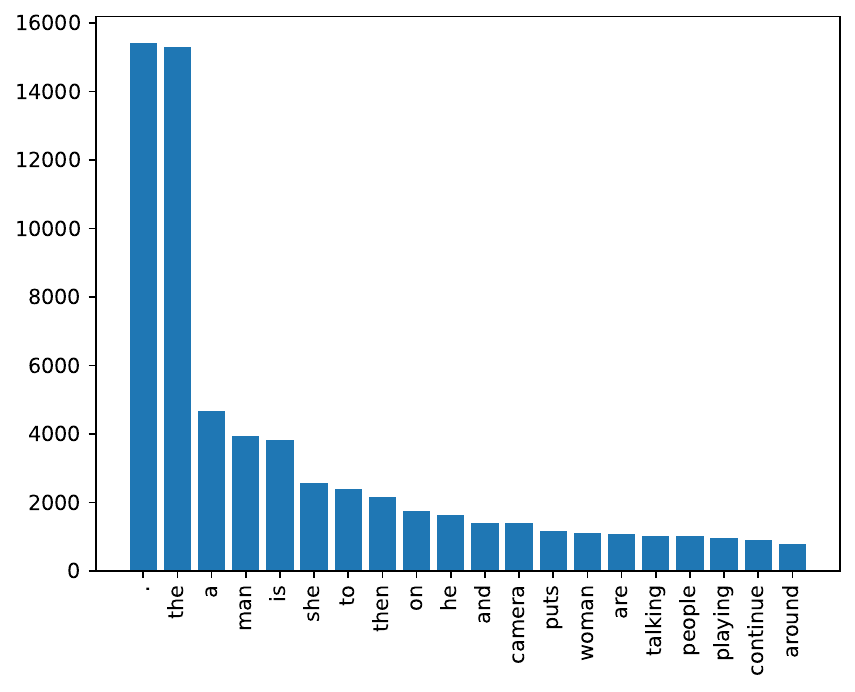}
    }
    \hfill
    \subcaptionbox{20 tokens \label{fig:long_freq}}{
        \includegraphics[width=0.3\linewidth]{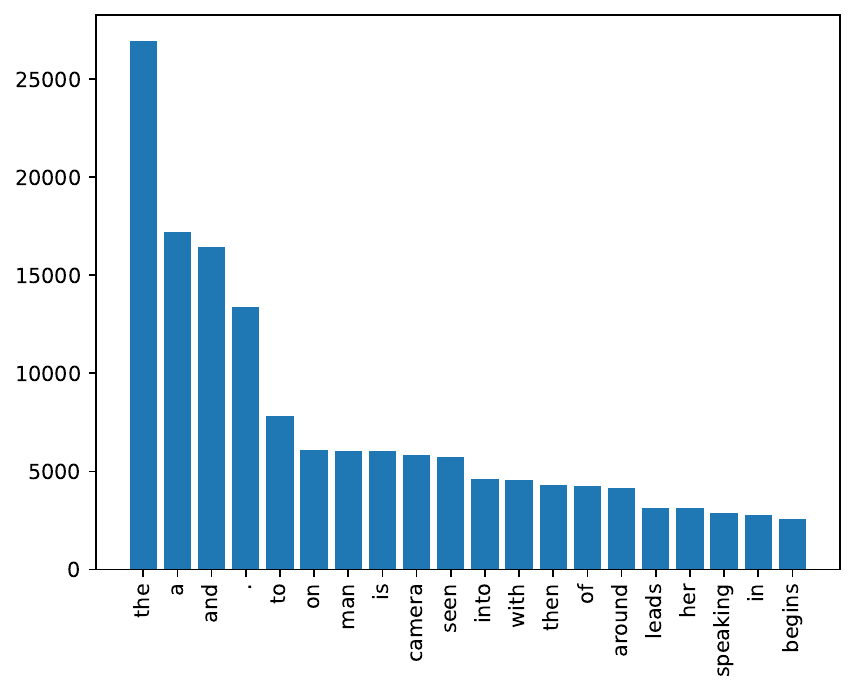}
    }

    \smallskip

    \subcaptionbox[t]{GT length \label{fig:spoken_mit_gt_freq}}{
        \includegraphics[width=0.3\linewidth]{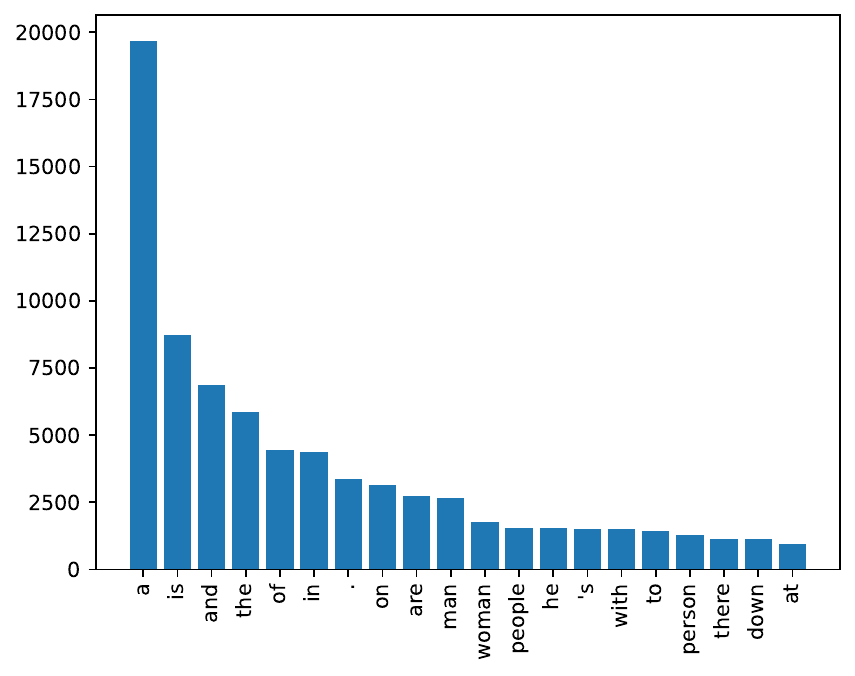}
    }
    \hfill
    \subcaptionbox[t]{5 tokens \label{fig:spoken_mit_min_freq}}{
        \includegraphics[width=0.3\linewidth]{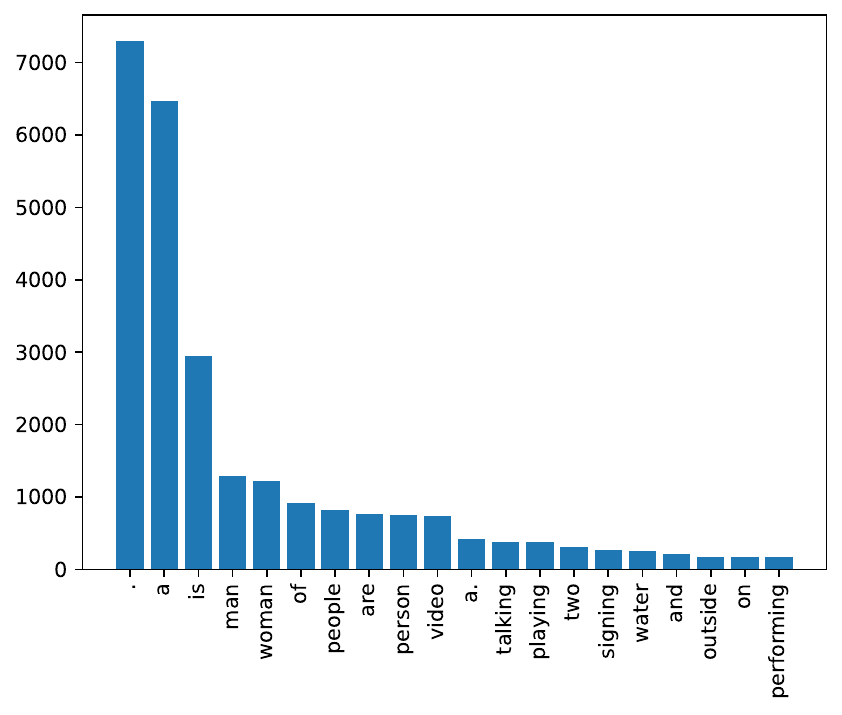}
    }
    \hfill
    \subcaptionbox[t]{20 tokens \label{fig:spoken_mit_long_freq}}{
        \includegraphics[width=0.3\linewidth]{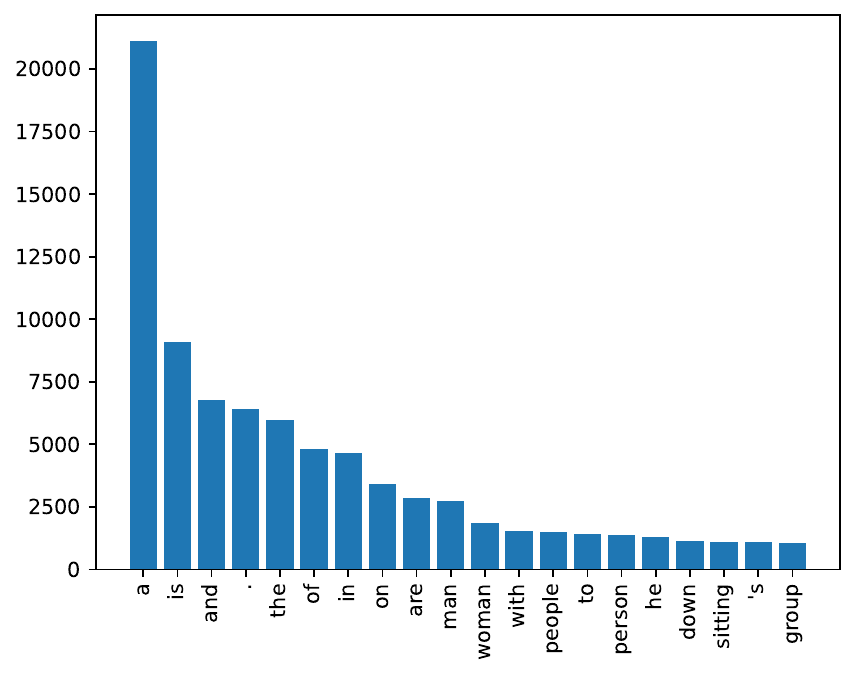}
    }

    \caption{
        Histograms of top 20 word tokens in generated captions for different target lengths
        trained on 
        \subref{fig:gt_freq}--\subref{fig:long_freq}
        ActivityNet Captions, and 
        \subref{fig:spoken_mit_gt_freq}--\subref{fig:spoken_mit_long_freq} 
        Spoken MiT.
        }
    \label{figs:freq_activity}
    \label{figs:freq_spoken}

\end{figure*}

The top 20 words with high frequency in the captions generated for ActivityNet Captions are shown in Figure~\ref{figs:freq_activity}\subref{fig:gt_freq}--\subref{fig:long_freq}.
The histogram for GT length shows that common articles and prepositions such as ``the'' and ``to'' are among the top 20 words. The period ``{.}'' also appears frequently, aligning closely with the number of sentences in the ActivityNet Captions validation set because every caption ends with a period. This implies that words with a higher frequency than ```{.}'' often appear more than once within a sentence. Given that ``the'' has nearly double the frequency of ``{.}'', it suggests that nouns are typically used twice in generated captions for GT length.
For the target length of 5, the frequency of ``and'' is seen to decrease; it occurs less than 2,000 times compared to over 10,000 times for GT length.  This suggests that limiting the length to 5 might result in fewer nouns, since ``and'' often combine nouns. The similar frequencies of ``the'' and ``{.}'' further support this, suggesting a decrease in the frequency of nouns.
However, for a target length of 20, the frequency of ``and'' and ``a'' becomes larger than ``{.}'', suggesting an increase in noun usage and redundancy to elongate the generated sentences.

The histograms for Spoken MiT are shown in 
Figure~\ref{figs:freq_spoken}\subref{fig:spoken_mit_gt_freq}--\subref{fig:spoken_mit_long_freq}.
We can see that articles and verbs are dominant for GT length, which is a similar trend to ActivityNet Captions. The difference is attributed to the presence of human-related nouns, such as ``people'', ``man'', and ``person''.
For a target length of 5, we can see a decrease in the frequency of prepositions, and verbs like ``playing'' and ``taking'' become more frequent than for the GT length. This might indicate that reasonable sentences are generated with a small set of nouns and verbs.
For a target length of 20, prepositions such as ``in'' and ``of'' and the article ``the'' becomes less frequent than periods, while ``and'' are more frequent than periods. This suggests that long sentences are generated by forming compound sentences with ``and''.

\subsection{Similarity between embeddings}

\begin{figure}[t]
    \centering
    
    \subcaptionbox{Length Level \label{fig:one-hot_vector}}{%
        \includegraphics[width=0.32\linewidth]{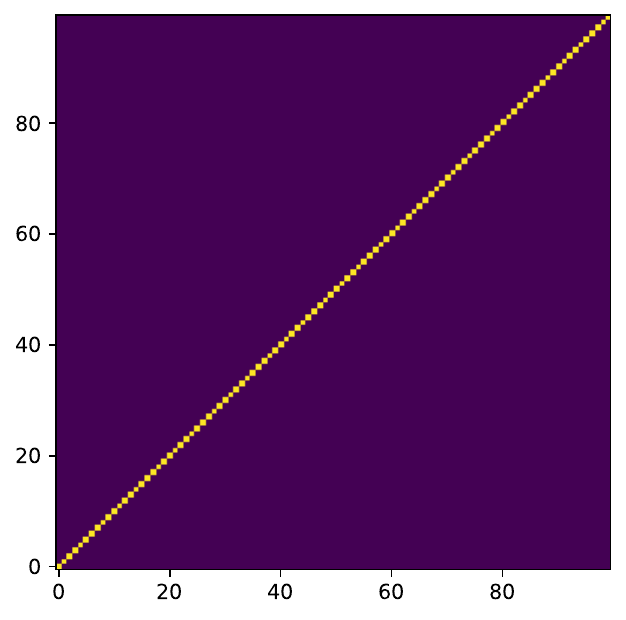}%
    }
    \hfill
    \subcaptionbox{Bit \label{fig:bit_vector}}{%
        \includegraphics[width=0.32\linewidth]{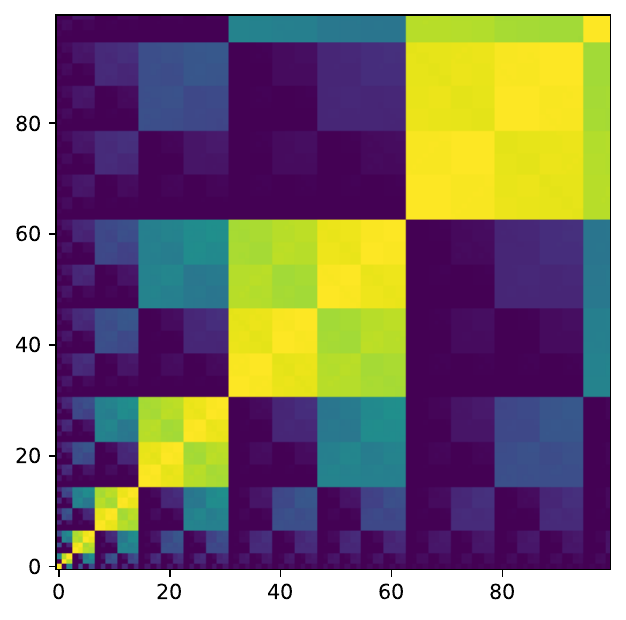}%
    }
    \hfill
    \subcaptionbox{Ordinal \label{fig:ordinal_vector}}{%
        \includegraphics[width=0.32\linewidth]{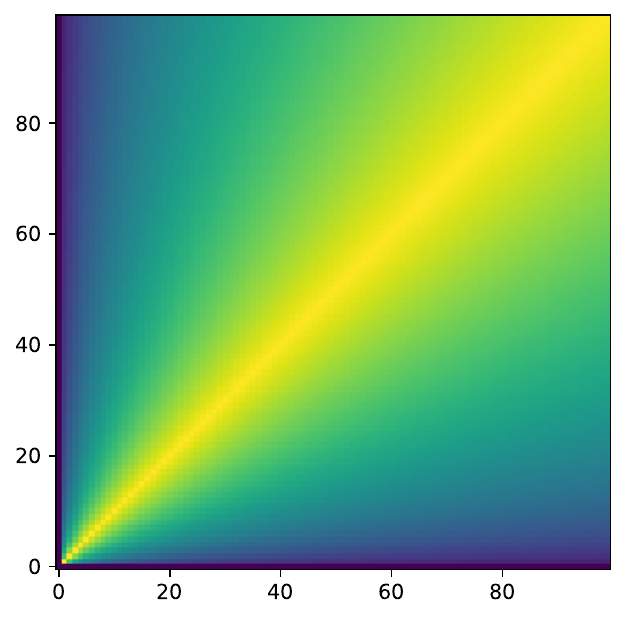}%
    }

    \smallskip

    \subcaptionbox{Length Level \label{fig:anetc_length_level_emb}}{%
        \includegraphics[width=0.32\linewidth]{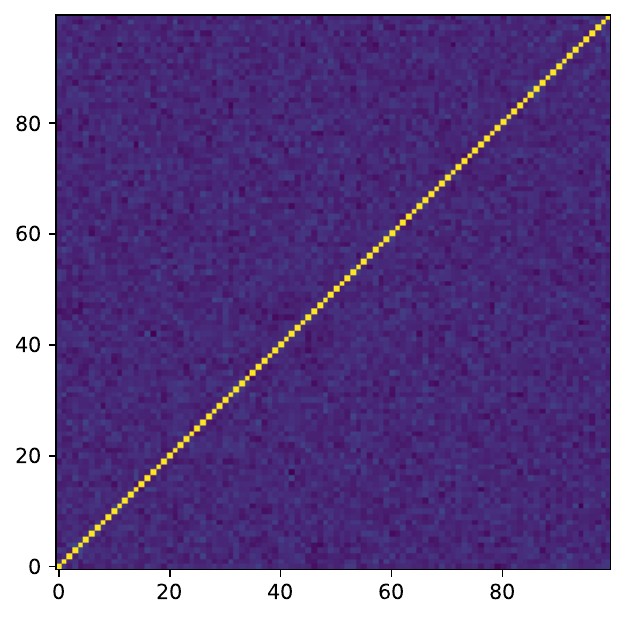}%
    }
    \hfill
    \subcaptionbox{Bit \label{fig:anetc_bit_emb}}{%
        \includegraphics[width=0.32\linewidth]{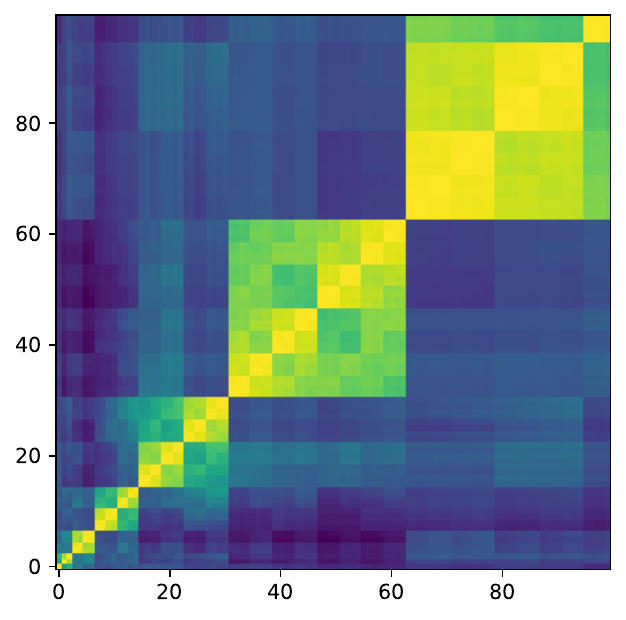}%
    }
    \hfill
    \subcaptionbox{Ordinal \label{fig:anetc_ordinal_emb}}{%
        \includegraphics[width=0.32\linewidth]{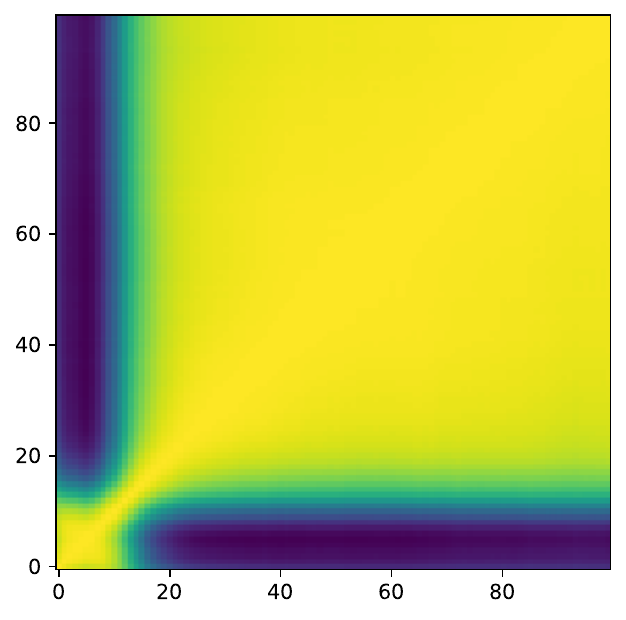}%
    }

    \smallskip

    \subcaptionbox{Length Level \label{fig:length_level_emb}}{%
        \includegraphics[width=0.32\linewidth]{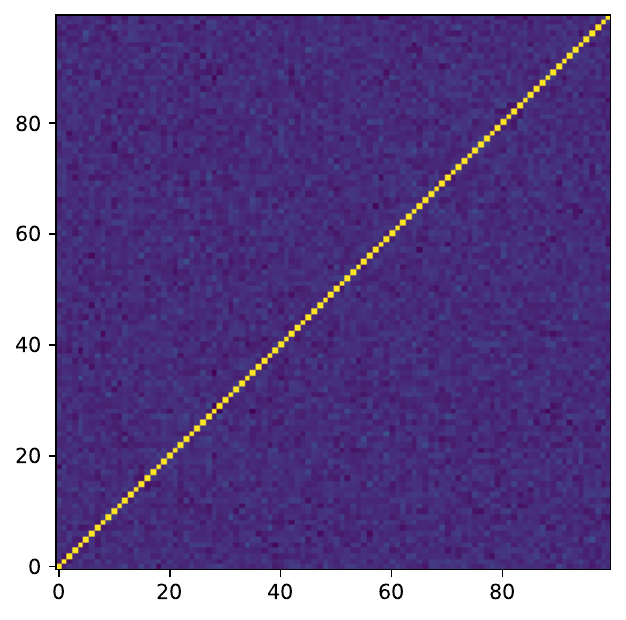}%
    }
    \hfill
    \subcaptionbox{Bit \label{fig:bit_emb}}{%
        \includegraphics[width=0.32\linewidth]{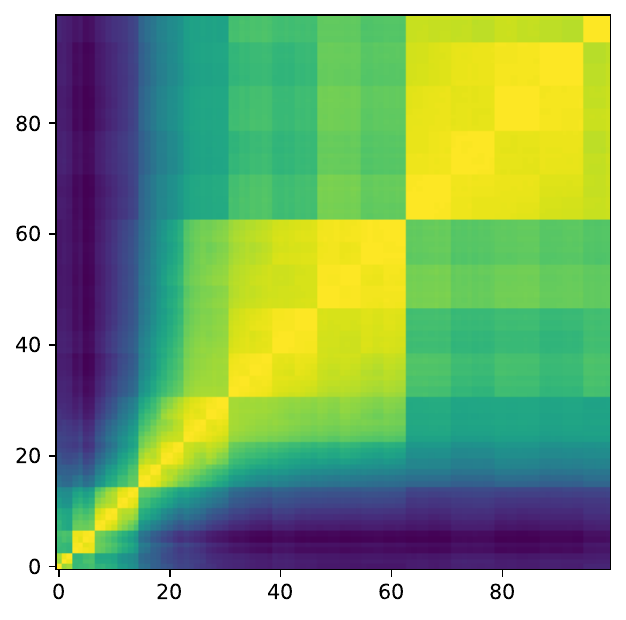}%
    }
    \hfill
    \subcaptionbox{Ordinal \label{fig:ordinal_emb}}{%
        \includegraphics[width=0.32\linewidth]{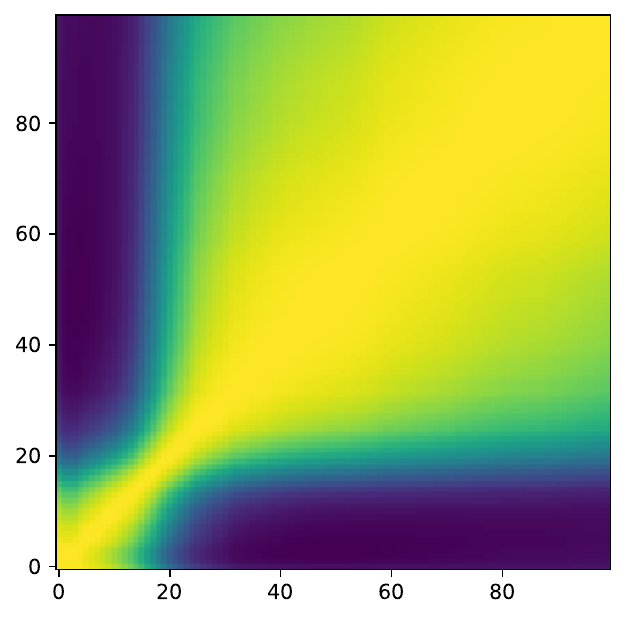}%
    }

  \caption{The similarity between
  \subref{fig:one-hot_vector}--\subref{fig:ordinal_vector}
  length representation vectors $\boldsymbol{t}_{k}$, and
  \subref{fig:anetc_length_level_emb}--\subref{fig:ordinal_emb}
  embedding vectors $\boldsymbol{e}_{k}$ for each type of embedding
  trained on (middle) ActivityNet Captions and (bottom) Spoken MiT.
  }
  \label{fig:vector_cosine}
  \label{figs:length_embed}

\end{figure}

As previously discussed, ordinal embedding effectively manages shorter lengths, even with limited data, but struggles with longer lengths. This trend also appears in learned embeddings. Figure \ref{fig:vector_cosine} shows the cosine similarity between embeddings of different lengths in the range [1, 101].

For length-level embeddings, even a small change in length results in low similarity, as expected.
This is because different lengths are not treated as ordinal numbers, but as distinct categories like word embeddings.
In bit embedding, notable changes in similarity occur around 16, 32, and 64, corresponding to changes in binary bits, although the change is less abrupt than in the length representation vector.

The similarity of ordinal embedding vectors changes smoothly. However, a trend change is noticeable around a length of 10 for ActivityNet Captios (Fig.\ref{fig:anetc_ordinal_emb}) and 20 for Spoken MiT (Fig.\ref{fig:ordinal_emb}); for shorter lengths, the similarity between embeddings is related to the difference in length, but for embeddings longer than those lengths, this relationship does not appear to hold. This could be due to the lack of data for longer captions, with captions of those lengths the most common, as in Figure \ref{fig:activitynet_hist}.

These findings indicate that the embedding representation significantly affects the trend in similarity between embeddings.


\subsection{Analysis Using ICA}

\begin{table}
    \centering
    \caption{Top 25 embeddings of word tokens and lengths that strongly respond to two independent components of dimensions 94 and 97.}
    \label{tab:ica}

\begin{tabular}{c|lr|lr}
rank & \multicolumn{2}{c|}{dim 94} & \multicolumn{2}{c}{dim 97} \\ \hline
1    & highway          & 100      & videos         & 100       \\
2    & high             & 100      & videos         & 30        \\
3    & throw            & 100      & videos         & 40        \\
4    & drawing          & 100      & videos         & 5         \\
5    & street           & 100      & videos         & 90        \\
6    & high             & 90       & videos         & 50        \\
7    & highway          & 90       & videos         & 20        \\
8    & basketball       & 100      & videos         & 80        \\
9    & stick            & 100      & videos         & 60        \\
0    & cross            & 100      & videos         & 70        \\
11   & foreign          & 100      & videos         & 10        \\
12   & throw            & 90       & video          & 100       \\
13   & how              & 100      & video          & 30        \\
14   & sign             & 100      & video          & 40        \\
15   & signing          & 100      & video          & 5         \\
16   & cut              & 100      & video          & 90        \\
17   & drawing          & 90       & video          & 50        \\
18   & Indian           & 100      & video          & 20        \\
19   & vehicle          & 100      & video          & 80        \\
20   & street           & 90       & video          & 60        \\
21   & soccer           & 100      & video          & 70        \\
22   & lights           & 100      & video          & 10        \\
23   & basketball       & 90       & film           & 100       \\
24   & shows            & 100      & pictures       & 100       \\
25   & stick            & 90       & film           & 30       
\end{tabular}

\end{table}

\begin{figure}[t]
    \centering

    \subcaptionbox{dim 94 \label{fig:length_box}}{%
        \includegraphics[width=0.48\linewidth]{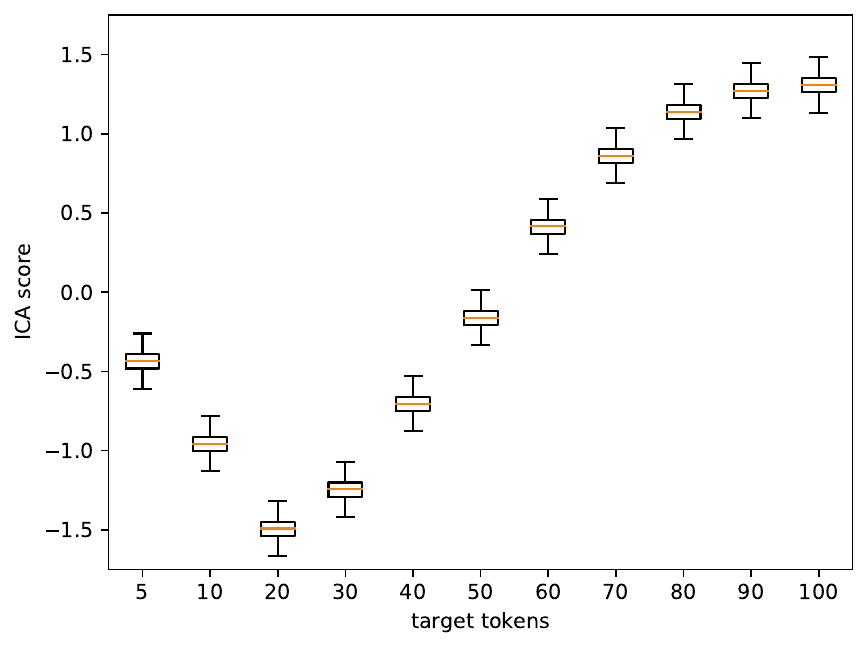}%
    }
    \hfill
    \subcaptionbox{dim 97 \label{fig:video_box}}{%
        \includegraphics[width=0.48\linewidth]{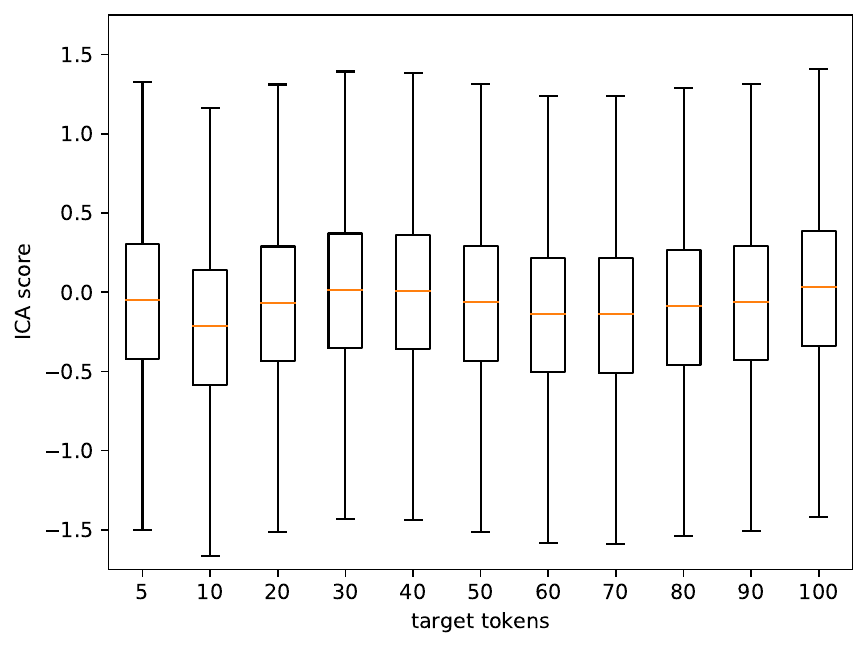}%
    }

    \smallskip

    \subcaptionbox{dim 94 \label{fig:length_94}}{%
        \includegraphics[width=0.48\linewidth]{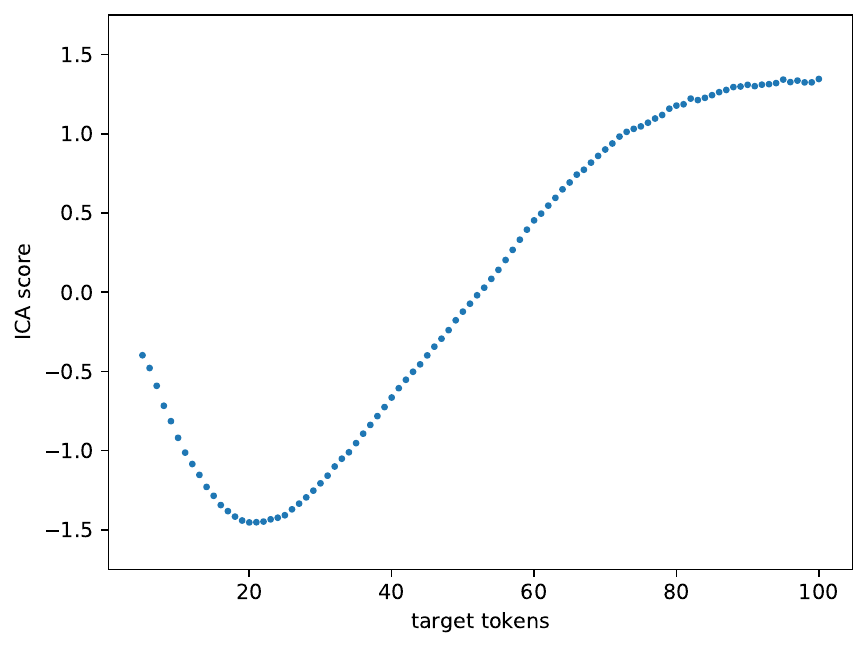}%
    }
    \hfill
    \subcaptionbox{dim 97 \label{fig:length_97}}{%
        \includegraphics[width=0.48\linewidth]{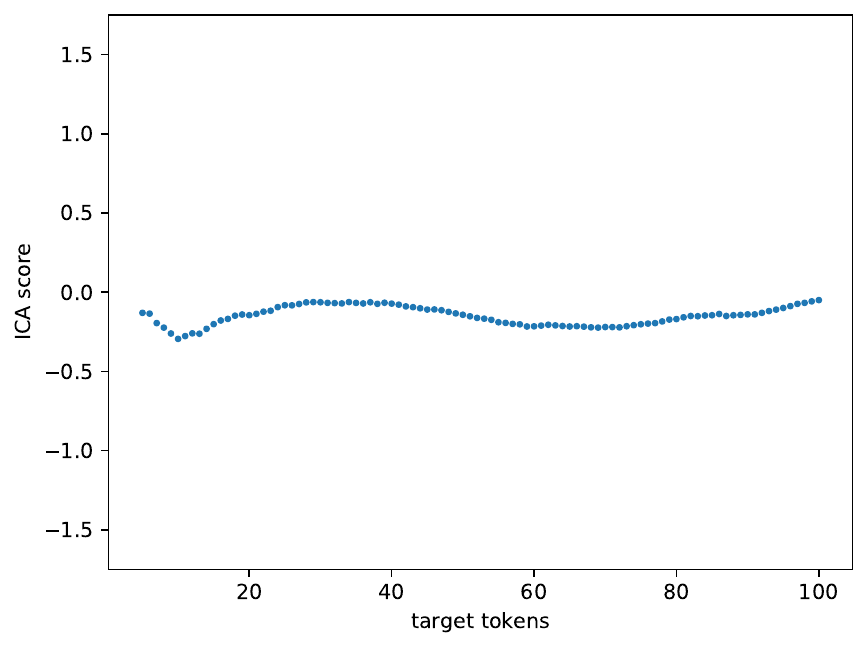}%
    }

  \caption{
      \subref{fig:length_box}\subref{fig:video_box}
      The distribution of word and length embedding responses in dimensions 94 and 97 for different lengths, and
      \subref{fig:length_94}\subref{fig:length_97}
      the same but for the length embedding only.
  }
  \label{figs:box}
  \label{figs:length}

\end{figure}

Independent Component Analysis (ICA) is useful for analyzing the semantics of word embedding components \cite{yamagiwa2023discovering}. Here, we used FastICA \cite{hyvarinen_fast_1999} to analyze word and length embeddings trained on the Spoken MiT dataset with ordinal embedding.
We analyzed the top 1000 frequent words in the dataset and 11 lengths ranging from 5 to 100. This resulted in $1000 \times 11$ embeddings, which we obtained by the addition
$\boldsymbol{W}^{\top}_{w} \boldsymbol{y}_{i} + \boldsymbol{e}_{l, k}$ as in Eq.\eqref{eq:length}. We then applied the ICA transformation to these embedding vectors and sorted the resulting independent components based on the absolute value of their kurtosis.

Table \ref{tab:ica} shows the embeddings that respond strongly to the independent components. This table includes the top 25 word tokens and lengths associated with the embedding vectors most strongly represented by dimensions 94 and 97, where we observed distinct characteristics of length embedding and word embedding.
In dimension 97, semantically similar words related to ``video'' exhibit a strong response regardless of length. In contrast, dimension 94 tends to have embeddings that strongly respond to length. The length of 100 is predominant, while the dimension shows a strong negative response to a length of 20.

Figure \ref{figs:box}\subref{fig:length_box}\subref{fig:video_box} 
illustrates the distribution of embedding responses in dimensions 94 and 97 for different lengths, and 
Figure \ref{figs:length}\subref{fig:length_94}\subref{fig:length_97} 
shows the plot of the values for the length embedding only.
In dimension 94, the values change smoothly and decrease until 20 while increasing within the range of $[20, 100]$, suggesting that the length of 20 is represented along this dimension (in the negative direction). However, dimension 97 does not show a significant difference in the distribution for different lengths, and the variations are primarily influenced by the word embedding. These results suggest that dimensions representing length information are separated from those representing word semantics.

\section{Conclusion}

This paper proposed length embedding methods for controllable video captioning. We introduced bit embedding and ordinal embedding through multi-hot length representation vectors. This enables us to control the length in tokens as well as the duration in seconds of the generated captions. Experimental results with two datasets showed that ordinal embedding controls the generated caption length most effectively. In addition, we presented an analysis of the generated captions and embedding vectors with ICA to illustrate how length is represented in the embeddings.
One of the future works is handling long captions. The current datasets do not contain long captions, so if the target length is 50 or more, the length of the generated captions becomes unstable. This requires either enhancing the datasets or applying data augmentation to learn from long captions.

\section*{\uppercase{Acknowledgements}}

This work was supported in part by JSPS KAKENHI Grant Number JP22K12090.

{
    \small
    \bibliographystyle{ieeenat_fullname}
    \bibliography{mybib,all,additional}
}

\appendix
\clearpage

\section{Additional examples}
\label{sec:appendix}

\subsection{Longer target lengths}

We have shown in Figures \ref{fig:activitynet_vid_cap} and \ref{fig:spoken_mit_vid_cap}
examples of generations with lengths of 5 or 20.
Here, we present examples generated up to a length of 100 using each embedding.

Figures
\ref{fig:5to100-length-level-anet},
\ref{fig:5to100-bit-embed-anet}, and
\ref{fig:5to100-ordinal-embed-anet}
show generated captions for the video of ActivityNet Captions
in Figure \ref{fig:activitynet_vid_cap}.
Length and bit embeddings generate strange sentences as the target length increases. However, with order embedding, reasonable sentences seem to be generated even as the length increases. When the target length is extremely short, such as 2, 3, or 4, none of the embedding methods succeeds in generating sentences of the target length because there are no captions of such lengths in the dataset.
As mentioned in section \ref{sec:Controlling for longer sentences}, the caption length generated by the ordinal embedding tends to shorten as the target length increases. However, with length-level embeddings, the length of the generated caption becomes unstable as the target length increases.

Figures
\ref{fig:5to100-length-level-smit},
\ref{fig:5to100-bit-embed-smit}, and
\ref{fig:5to100-ordinal-embed-smit}
show generated captions for the video of Spoken MiT
in Figure \ref{fig:spoken_mit_vid_cap}.
Since this dataset contains many short captions, the generated sentences tend to be short regardless of the embedding method used, even if the target value is set extremely short. However, with length-level embeddings, sentences with repetitive words are generated when the target length increases. This trend is also observed with both bit and order embeddings, but they produce more coherent sentences in comparison to length-level embeddings.

\begin{figure*}[t]
    \centering

    {\small
    \begin{tabular}{ccp{.8\linewidth}}
    \multicolumn{2}{c}{length} & \multirow{2}{*}{captions with length-level embedding} \\
      target & gen &\\ \hline
2 & 11 &  The then stops back on back on starts back back back \\
3 & 12 &  The The men is standing in standing in standing and he is \\
4 & 13 &  The man then sitting in black black shirt's black shirt is standing \\
5 & 15 &  The man then the man is pulling the straps of the pedal and he's \\
6 & 7 &  He then adjusts the lever.......... \\
7 & 7 &  He then adjusts the bike............ \\
8 & 8 &  He then adjusts the lever lever............ \\
9 & 9 &  He adjusts the lever and adjusts it........... \\
10 & 11 &  The man adjusts the handle lever and adjusts it........... \\
11 & 12 &  The man pulls the lever lever and adjusts the lever.......... \\
12 & 13 &  The man adjusts the settings of the rowing machine.. too........ \\
13 & 14 &  The man adjusts the lever and adjusts the pedal lever. the handle......... \\
14 & 15 &  The man pulls the handle lever and adjusts the cables to the bike.......... \\
15 & 22 &  The man adjusts the lever and adjusts the lever to the pedal.The handle lever. The handle. The handle \\
16 & 23 &  The man then pulls the handlebars of the bike and adjusts the settings.The lever. the lever.
. \\
17 & 22 &  The man then pulls the handlebars up and begins to pedal the pedals. the handle.
.
.... \\
18 & 21 &  The man then pulls the handlebars up and down on the rowing machine. the lever.
....... \\
19 & 26 &  The man then pulls the handle lever up and down on the machine to raise the settings.The lever are positioned. The lever. \\
20 & 26 &  The man then pulls the bike up and down and begins working out of the pedal.The lever. the lever. the bike... \\
30 & 40 &  The man then pulls the handle lever and adjusts the settings of the rowing machine and adjusts the settings of the rowing machine and the cables. and adjusts the cables. and adjusts the cables. and \\
40 & 50 &  The man then puts the pedals back on the pedal bar and begins to pull the bar back and pull it back to the back, then he puts the bar back to the back. then he lowers the bar. then back. back. back. back \\
50 & 59 &  The man then pulls a dumb up and then put the pedals up and then put the cables up and put the cables up and put the cables up and put the cables up and put the back of the front of the front of the front of the front of the front of the bottom. the straps \\
60 & 70 &  The man then the black weight button is the same button of the weight of the weight of the weight of the weight of the weight of the weight of the weight of the handle, then the handle, then the handle, then the handle the handle, the handle, the handle, the handle, the handle, the handle the handle, the handle, \\
70 & 80 &  The man then sits to the pedals and the man is standing on the pedal and he's standing on the pedal and he's standing on the pedal and he's done the lever and he's done the lever and he's done the lever and he's done the lever and he's done the straps and he's done the straps and the straps and the straps and the straps and the straps and the straps \\
80 & 42 &  The man then pulls the bike on the ro and starts pulls the handle and starts the pedals to the machine and the pedal........................................... again..... again.. again. again.. again.. again. again.. again \\
90 & 59 &  The man then pulls the handle of lever of the weight of lever of the man then puts the lever and puts the lever of the lever of the lever of the lever of the lever of the straps of the straps.. the straps. the straps. the straps. the straps. the straps........................................ \\
100 & 110 &  The man man man man man is standing a weight bike bike bike bike bike bike and standing a weight weight weight and sitting a weight, and sitting a weight bike bike and sitting a weight bike. the exercise. the exercise. the exercise. the lever. the lever. the lever. the lever. the lever. the machine. the machine. the machine. the machine. the machine. the machine. the machine. the machine. the machine. the machine. the machine. the machine. the machine. the machine. the machine. the machine \\
\end{tabular}
    }

    \caption{Captions generated for the same video of ActivityNet Captions in Fig.\ref{fig:activitynet_vid_cap} with length-level embedding.
    The target length ranges from 2 and 100 tokens.}
    \label{fig:5to100-length-level-anet}

\end{figure*}

\begin{figure*}[t]
    \centering

    {\small
    \begin{tabular}{ccp{.8\linewidth}}
    \multicolumn{2}{c}{length} & \multirow{2}{*}{captions with bit embedding} \\
      target & gen &\\ \hline
2 & 11 &  The man then begins to use the machine on the machine \\
3 & 12 &  The man then begins to use the machine on the machine again \\
4 & 10 &  The man then adjusts the straps on the machine. \\
5 & 10 &  The man then adjusts the straps on the machine. \\
6 & 6 &  The man adjusts the straps. \\
7 & 7 &  The man then adjusts the bike. \\
8 & 7 &  The man then adjusts the straps. \\
9 & 10 &  The man then adjusts the straps on the machine. \\
10 & 10 &  The man then begins working out on the machine. \\
11 & 10 &  The man then begins working out on the machine. \\
12 & 11 &  The man then begins working out on the machine again. \\
13 & 13 &  The man then begins working out on the machine and adjusting it. \\
14 & 13 &  The man then begins to use the machine on the machine again. \\
15 & 15 &  The man then begins to use the machine on the machine and adjusting it. \\
16 & 16 &  The man then begins to use the machine on the machine and adjusts the settings. \\
17 & 16 &  The man then begins to use the machine on the machine and begins to work. \\
18 & 17 &  The man then begins to use the machine on the machine and begins to work out. \\
19 & 19 &  The man then pulls the handlebars back and begins to pull the handlebars back and forth. \\
20 & 21 &  The man then pulls the handlebars of the bike and begins to pull the handlebars back and forth. \\
30 & 31 &  The man then pulls the handlebars back and begins to pull the handlebars back and the man continues to talk to the camera and show the finished product. \\
40 & 35 &  The man then begins to pull the bar back and forth on the machine and begins to use the machine again and ends by looking back to the camera and smiling to the camera. \\
50 & 43 &  The man then begins to pull the bar back and forth on the machine and begins to pull the bar back and forth on the machine while the man continues to speak to the camera and the camera shows the machine again. \\
60 & 49 &  The man then begins to pull the bar back and forth on the machine and begins to pull the bar back and forth on the machine while the man continues to speak to the camera and the camera shows the machine's settings and the man's feet. \\
70 & 80 &  The man is using the back of the back of the back of the back of the back of the back of the back of the back of the back of the back of the back of the back of the back of the back of the back of the back of the back of the back of the back of the back of the back of the back of the back of the back of the back of the \\
80 & 90 &  The man is sitting on the handle of the handle of the handle of the handle of the handle of the handle of the the the the the the the the the the the the the the the the the the the the the the the the the the the the the the the the the the the the the the the the the the the the the the the the the the the the the the the the the the the the the the the the the the the \\
90 & 100 &  The man is on the handle of the handle of the handle of the handle of the the the the the the the the the the the the the the the the the the the the the the the the the the the the the the the the the the the the the the the the the the the the the the the the the the the the the the the the the the the the the the the the the the the the the the the the the the the the the the the the the the the the \\
100 & 110 &  The weight the weight the weight the weight the weight the weight the weight the weight the weight the weight the weight the weight the weight the weight the weight the weight the weight the weight the weight the weight the weight the weight weight the weight the weight the weight the weight the weight the weight. the weight the weight the weight. the weight. the weight. the weight the weight. the weight. the weight the weight the weight the weight the weight the weight the weight the weight the weight the weight the weight the weight the weight the weight the weight the \\
      \end{tabular}
    }

    \caption{Captions generated for the same video of ActivityNet Captions in Fig.\ref{fig:activitynet_vid_cap} with bit embedding.
    The target length ranges from 2 and 100 tokens.}
    \label{fig:5to100-bit-embed-anet}

\end{figure*}

\begin{figure*}[t]
    \centering

    {\small
    \begin{tabular}{ccp{.8\linewidth}}
    \multicolumn{2}{c}{length} & \multirow{2}{*}{captions with ordinal embedding} \\
      target & gen &\\ \hline
2 & 9 &  The man is working out on the machine. \\
3 & 9 &  The man is working out on the machine. \\
4 & 9 &  The man is working out on the machine. \\
5 & 6 &  The man is working out. \\
6 & 6 &  The man is working out. \\
7 & 6 &  The man is working out. \\
8 & 9 &  The man is working out on the machine. \\
9 & 9 &  The man is working out on the machine. \\
10 & 9 &  The man then adjusts the bike's settings. \\
11 & 10 &  The man then begins working out on the machine. \\
12 & 10 &  The man then begins working out on the machine. \\
13 & 11 &  The man then begins to work out on the machine. \\
14 & 14 &  The man then begins to work out on the machine and adjusting it. \\
15 & 14 &  The man then begins to work out on the machine and adjusts it. \\
16 & 15 &  The man then begins to work out on the machine and begins to exercise. \\
17 & 17 &  The man then begins to work out on the machine and begins to pull himself back. \\
18 & 19 &  The man then begins to work out on the machine and begins to pull himself back and fourth. \\
19 & 19 &  The man then begins to work out on the machine and begins to pull himself back and fourth. \\
20 & 21 &  The man then begins to work out on the machine and begins to pull himself back and fourth on it. \\
30 & 28 &  The man then begins to work out on the machine and begins to pull himself up and down on the machine while still speaking to the camera. \\
40 & 39 &  The man then begins to work out on the machine and begins to pull the weight back and fourth while still speaking to the camera and showing off his body and the machine's settings on the machine. \\
50 & 55 & A man is seen sitting on a piece of exercise equipment and begins moving himself back and fourth on the machine while another man is seen sitting on the machine and the man begins moving himself back and fourth on the machine and the man begins moving himself back and fourth on the machine. \\
60 & 56 & A man is seen sitting on a piece of exercise equipment and begins moving himself back and fourth on the machine and begins moving his body back and fourth while looking back to the camera and speaking to the camera and pointing to the machine and the man's feet and the machine's settings. \\
70 & 56 & A man is seen sitting on a piece of exercise equipment and begins moving himself back and fourth on the machine and begins moving his body back and fourth while looking back to the camera and speaking to the camera and pointing to the machine and the man's feet and the machine's settings. \\
80 & 55 & A man is seen sitting on a piece of exercise equipment and begins moving himself back and fourth on the machine while another man is seen sitting on the machine and the man begins moving himself back and fourth on the machine and the man begins moving himself back and fourth on the machine. \\
90 & 59 & A man is seen sitting on a piece of exercise equipment and begins moving himself back and fourth on the machine and begins moving his body back and fourth while the man continues to push himself back and fourth on the machine and ends with the man sitting on the machine and the man sitting on the machine. \\
100 & 54 & A man is seen sitting on a piece of exercise equipment and begins moving himself back and fourth on the machine while another man is seen sitting on the other side and the man begins moving himself back and fourth on the machine and begins moving himself back and fourth on the machine. \\
    \end{tabular}
    }

    \caption{Captions generated for the same video of ActivityNet Captions in Fig.\ref{fig:activitynet_vid_cap} with ordinal embedding.
    The target length ranges from 2 and 100 tokens.}
    \label{fig:5to100-ordinal-embed-anet}

\end{figure*}

\begin{figure*}[t]
    \centering

    {\small
    \begin{tabular}{ccp{.8\linewidth}}
    \multicolumn{2}{c}{length} & \multirow{2}{*}{captions with length-level embedding} \\
      target & gen &\\ \hline
2 & 7 & a baby video of a baby playing \\
3 & 3 & a baby. \\
4 & 8 & a baby is playing with the water. \\
5 & 3 & a baby. \\
6 & 6 & a baby playing with toys. \\
7 & 8 & a baby playing in a tub........... \\
8 & 9 & a baby is playing in a tub........... \\
9 & 9 & a baby is playing with a toy car......... \\
10 & 9 & a baby is playing with a toy car........ \\
11 & 12 & a baby is playing with a toy in the sink........... \\
12 & 12 & a baby is playing with a toy that is yellow in..... \\
13 & 14 & a baby is playing with a toy that is in the bath........... \\
14 & 15 & a baby is playing with a toy that is in a small bucket........... \\
15 & 15 & a baby is sitting in a bathtub with a yellow sponge in it...... \\
16 & 16 & a baby is sitting in a bathtub with a yellow sponge in her hand.... \\
17 & 17 & a baby is sitting in a bathtub with a yellow sponge in front of it........ \\
18 & 17 & a baby is sitting in a bathtub with a yellow sponge and a yellow sponge. \\
19 & 19 & a baby is sitting in a bathtub with a yellow sponge in front of her and playing....... \\
20 & 20 & a baby is sitting in a bathtub with a yellow sponge in front of her playing with it..... \\
30 & 31 & a baby is sitting in a bathtub with a yellow sponge in front of it and a yellow sponge in the background and a woman is talking. Screen. \\
40 & 44 & a baby is sitting in a bathtub with a yellow sponge in front of it and a yellow sponge in the bathtub and the baby is playing with the sponge and the sponge is yellow and white. Com. Com.. \\
50 & 60 & a baby is sitting in a bathtub with a yellow sponge in front of it and a yellow sponge in the background and a woman is standing over the baby and she is putting the sponge in the baby's mouth and she is smiling at the baby. She is smiling. She is smiling. She is \\
60 & 67 & a baby is sitting in a bathtub with a yellow sponge in front of it and a yellow sponge in the background and a baby is sitting in the bathtub with a yellow sponge in front of it and a baby is playing with the sponge and the baby is smiling and giggling. There is no sound.. Video. Video. Video \\
70 & 73 & a baby is sitting in a bathtub with a yellow plastic bucket in front of it and a yellow bucket in the background and there is a woman who is holding a yellow sponge and she is pouring water into the bucket and the baby is looking at the bucket and she is smiling and she is holding the bucket in her hand. The baby is playing with the water........ \\
80 & 90 & a baby is sitting in a bathtub with a yellow sponge in it and a yellow sponge in the background and there is a woman who is sitting in the bathtub and she is holding a yellow sponge in her hand and she is putting it into the bucket and she is putting it into the bucket and she is putting it into the bucket and she is smiling and she is talking to the baby. She is smiling. She is smiling. She is smiling \\
90 & 89 & a baby is sitting in a bathtub with a yellow plastic bucket in front of it and there is a yellow plastic bucket in the background and there is a little girl who is sitting in the bathtub and she is playing with the water and she is playing with the water and she is playing with the water and she is smiling and she is enjoying the water and she is enjoying the water and she is enjoying the water and enjoying the water........... \\
100 & 110 & a baby is sitting in a red bath and a bathtub and a bathtub and there is a yellow and yellow toy that is sitting on the floor and there is a yellow and yellow and yellow and yellow and yellow and yellow and yellow and yellow and yellow and yellow and yellow and yellow and yellow and yellow and yellow and yellow and yellow and yellow and yellow and yellow and yellow and white and blue and white and blue and white and blue and white and white and blue and white and white and blue and white and white and white and white and blue and \\
      \end{tabular}
    }

    \caption{Captions generated for the same video of Spoken MiT in Fig.\ref{fig:spoken_mit_vid_cap} with length-level embedding.
    The target length ranges from 2 and 100 tokens.}
    \label{fig:5to100-length-level-smit}

\end{figure*}

\begin{figure*}[t]
    \centering

    {\small
    \begin{tabular}{ccp{.8\linewidth}}
    \multicolumn{2}{c}{length} & \multirow{2}{*}{captions with bit embedding} \\
      target & gen &\\ \hline
2 & 11 & a baby is playing in a small plastic tub with bubbles \\
3 & 12 & a baby is playing in a small plastic tub with bubbles. \\
4 & 7 & a baby playing in a tub.. \\
5 & 8 & a baby is playing in a tub... \\
6 & 6 & a baby playing in bubbles....... \\
7 & 7 & a baby playing in a tub........ \\
8 & 8 & a baby is playing in a tub......... \\
9 & 9 & a baby is playing in a small pool........ \\
10 & 10 & a baby is playing in a small plastic tub......... \\
11 & 11 & a baby is playing in a small plastic water fountain....... \\
12 & 12 & a baby is playing in a small tub with a hose...... \\
13 & 13 & a baby is playing in a small tub with a water pitcher...... \\
14 & 14 & a baby is playing in a small plastic tub with a water pitcher..... \\
15 & 15 & a baby is playing in a small plastic tub with a water fauc. \\
16 & 16 & a baby is playing in a small plastic tub with a hose in her mouth. \\
17 & 17 & a baby is sitting in a red plastic tub with a yellow sponge and a bucket. \\
18 & 18 & a baby is sitting in a red plastic tub with a yellow sponge and a pink bucket.... \\
19 & 19 & a baby is sitting in a red plastic tub with a yellow sponge and a pink plastic bucket..... \\
20 & 20 & a baby is sitting in a red plastic tub with a yellow sponge and a pink bucket and smiling. \\
30 & 29 & a baby is sitting in a red plastic tub with a pink bucket in it and a hand comes in and sprays the baby with a hose. \\
40 & 40 & a little baby is sitting in a red plastic tub with a pink top on and a yellow bottom and she's holding a water pitcher and she's pouring water on top of the baby's wet hair. \\
50 & 53 & a little baby is sitting in a red plastic tub with a pink top on it and a yellow bottom and she's holding a pink plastic spoon in her hand and she's putting water on top of the baby and she's talking to the baby. Baby. Baby. \\
60 & 61 & there's a little baby sitting in a red plastic tub with a red plastic top on it and there's a little girl with a yellow plastic cup in her hand and she's pouring water on the baby and the baby is looking at the water and she's smiling and laughing and smiling at the same time. \\
70 & 80 & there's a little baby sitting in a red plastic tub with a red plastic top on it and there's a little girl with a yellow plastic cup in her hand and she's pouring water on the baby and the baby is looking at the water and she's smiling and laughing and the baby is looking at the water and she's smiling at the baby. She's smiling. She's smiling. She's \\
80 & 81 & there's a little baby sitting in a red plastic tub with a red plastic top on it and there's a little girl with a yellow plastic spoon in her hand and she's pouring water on the baby and the baby is looking at the water and she's smiling and laughing and the baby is looking at the water and she's smiling and laughing and the baby is looking at the water and smiling and laughing. \\
90 & 100 & there's a little baby sitting in a red plastic tub with a red top and a white bottom and there's a little girl sitting in the red tub and she's got a pink top and she's got a little girl in the red tub and she's got a little girl in the yellow tub and she's got a little girl in the red tub and she's putting water on the baby and the baby is smiling and laughing and she's smiling and smiling and smiling. She's happy. She's \\
100 & 110 & there's a little baby sitting in a red plastic tub with a red top and a white bottom and there's a little girl sitting in the red tub and she's holding a little baby in her hands and she's pouring water on the baby and the baby is looking at the water and the little girl is smiling and the baby is looking at the water and the little girl is smiling and the baby is looking at the water and the little girl is smiling and the baby is looking at the water and the baby is smiling and the baby is smiling and the \\
      \end{tabular}
    }

    \caption{Captions generated for the same video of Spoken MiT in Fig.\ref{fig:spoken_mit_vid_cap} with bit embedding.
    The target length ranges from 2 and 100 tokens.}
    \label{fig:5to100-bit-embed-smit}

\end{figure*}

\begin{figure*}[t]
    \centering

    {\small
    \begin{tabular}{ccp{.8\linewidth}}
    \multicolumn{2}{c}{length} & \multirow{2}{*}{captions with ordinal embedding} \\
      target & gen &\\ \hline
      2 & 7 & a baby is playing in bubbles. \\
      3 & 5 & a baby is playing. \\
      4 & 5 & a baby is playing. \\
      5 & 5 & a baby is playing. \\
      6 & 6 & a baby playing in water. \\
      7 & 7 & a baby is playing with bubbles. \\
      8 & 8 & a baby is playing in a pool. \\ 
      9 & 9 & a baby is playing in a small pool. \\ 
      10 & 10 & a baby is taking a bath in a tub. \\
      11 & 11 & a baby is taking a bath in a small tub. \\ 
      12 & 12 & a baby is playing in a small tub with a toy. \\ 
      13 & 13 & a baby is playing in a small tub with a yellow sponge. \\ 
      14 & 14 & a baby is playing in a small tub with a yellow rubber object. \\ 
      15 & 15 & a baby is playing in a small tub with a yellow sponge and water. \\ 
      16 & 16 & a baby is sitting in a red plastic tub the baby is taking a bath. \\ 
      17 & 17 & a baby is sitting in a red plastic tub the baby is taking a bubble bath. \\ 
      18 & 18 & a baby is sitting in a red plastic tub the baby is taking a bath with bubbles. \\ 
      19 & 19 & a baby is sitting in a red plastic tub the baby is taking a bubble bath with bubbles. \\
      20 & 20 & a baby is sitting in a red plastic tub with water and a pink sponge the baby is laughing. \\
30 & 30 & a baby is sitting in a red plastic tub with a yellow sponge in it the baby is playing with a yellow sponge and a woman is standing nearby. \\
40 & 40 & a baby is sitting in a red plastic tub with a water hose in it and a sponge in the background the baby is wearing a diaper and there is a pink flower in the tub with a flower. \\
50 & 50 & there's a baby in a red plastic tub with a little bit of water in it and there's a little girl in a pink bikini sitting in the tub and she's taking a little water and putting it into the baby's mouth and then laughing. \\
60 & 60 & there's a baby in a red plastic tub with a little bit of water in it and there's a little girl in a pink bikini sitting in the tub and she's taking a little water bottle and she's taking it and putting it in the baby's mouth and she's smiling at the baby. \\
70 & 71 & there's a baby in a red plastic tub with a little bit of water in it and there's a little girl in a pink bikini top and she's sitting in the tub and she's taking a little water bottle and she's taking it and she's taking it out of the tub and she's taking it out of the tub and she's smiling. \\
80 & 87 & there's a baby in a red plastic tub with a little bit of water in it and there's a little girl in a pink bikini top and she's sitting in the tub and she's taking a little water bottle and she's taking it and she's taking it out of the tub and she's taking it out of the tub and she's taking it out of the tub and she's smiling at the baby. She's happy. \\
90 & 99 & there's a baby in a red plastic tub with a little bit of water in it and there's a little girl in a red plastic tub with a little bit of water in it and she's taking a little bit of water and she's taking a little bit of water and she's taking a little bit of water and she's taking a little water and she's taking a little water and she's taking a little water and she's taking a little water and she's taking a little water. \\
100 & 110 & there's a baby in a red plastic tub with a little bit of water in it and there's a little girl in the middle of the tub and she's sitting in the tub and she's got a sponge in her hand and she's taking a little bit of water and she's taking a little bit of water and she's taking a little bit of water and she's taking a little water and she's taking a little water and she's taking a little water and she's taking a little water and she's taking a little water and she's \\
      \end{tabular}
    }

    \caption{Captions generated for the same video of Spoken MiT in Fig.\ref{fig:spoken_mit_vid_cap} with ordinal embedding.
    The target length ranges from 2 and 100 tokens.}
    \label{fig:5to100-ordinal-embed-smit}

\end{figure*}

\clearpage
\subsection{Longer temporal durations with ordinal embedding}

Figure
\ref{fig:5to100-ordinal-embed-duration-control},
shows the duration-controlled captions generated
up to a duration of 10 seconds using the ordinal embedding
for the video of Spoken MiT
in Figure \ref{fig:spoken_mit_vid_cap}.
Even with 5 tokens, the duration exceeds one second. Therefore, even if the target duration is set lower, the generated sentences tend to be relatively long due to the lack of shorter captions in the dataset. 
When the target duration is increased to 10 seconds, the duration of the generated captions is fairly close to the target duration.
Note that increasing the target duration beyond 10 seconds will not change the output captions because the number of generated tokens is limited to 50 in the experiment.

\begin{figure*}[t]
    \centering

    {\small
    \begin{tabular}{ccp{.8\linewidth}}
    \multicolumn{2}{c}{duration} & \multirow{2}{*}{captions with ordinal embedding} \\
      target & gen &\\ \hline
0.2 & 3.099 & a little baby is sitting in a bathtub with a yellow sponge. (14 tokens) \\
0.3 & 2.620 & a little baby is playing with a toy in a red tub. (13 tokens) \\
0.4 & 2.191 & a little baby is playing in a bathtub. (10 tokens) \\
0.5 & 2.191 & a little baby is playing in a bathtub. (10 tokens) \\
1.0 & 1.035 & a little baby.................................. (5 tokens) \\
1.5 & 1.523 & a little baby is in a bath............................ (9 tokens) \\
2.0 & 2.095 & a little baby is playing in a bathtub........... (11 tokens) \\
2.5 & 2.656 & a little baby is playing with a toy in a bathtub. (13 tokens) \\
3.0 & 3.099 & a little baby is sitting in a bathtub with a yellow sponge. (14 tokens) \\
3.5 & 3.560 & a little baby is sitting in a red chair and is being given a bath. (16 tokens) \\
4.0 & 3.991 & a little baby is sitting in a red chair and a mother is pouring water on her. (18 tokens) \\
4.5 & 4.565 & a little baby is sitting in a red chair with a yellow sponge and a yellow sponge in her mouth. (21 tokens) \\
5.0 & 4.841 & a little baby is sitting in a red chair with a yellow sponge and a little yellow sponge in her mouth. (22 tokens) \\
6.0 & 5.873 & a little baby is sitting in a red chair with a yellow sponge in front of it and a little girl is sitting in a pink chair. (28 tokens) \\
6.5 & 6.765 & a little baby is sitting in a red chair with a yellow sponge in front of it and a little girl is sitting in a pink chair with a yellow sponge. (32 tokens) \\
7.0 & 7.166 & a little baby is sitting in a red chair with a yellow sponge in front of it and a little girl is sitting in a pink chair with a yellow sponge in front. (34 tokens) \\
7.5 & 8.469 & a little baby is sitting in a red chair with a yellow sponge in front of it and a little girl is sitting in a pink chair with a yellow sponge in front of her. She's laughing. (40 tokens) \\
8.0 & 8.141 & a little baby is sitting in a red chair with a yellow sponge in front of it and a little girl is sitting in a pink chair with a yellow sponge and she's washing her face. (38 tokens) \\
8.5 & 9.074 & a little baby is sitting in a red bathtub with a yellow sponge in it and a little girl is sitting in the bathtub with a yellow sponge and a little boy is sitting in the bathtub with a sponge. (44 tokens) \\
9.0 & 9.366 & a little baby is sitting in a red bathtub with a yellow sponge in it and a little girl is sitting in the bathtub with a yellow sponge and a little boy is sitting in the bathtub with a yellow sponge. (45 tokens) \\
9.5 & 10.262 & a little baby is sitting in a red bathtub with a yellow sponge in it and a little girl is sitting in the bathtub with a yellow sponge and a little boy is sitting in the bathtub with a yellow sponge and a little boy. (49 tokens) \\
10.0 & 10.377 & a little baby is sitting in a red bathtub with a yellow sponge in it and a little girl is sitting in the bathtub with a yellow sponge and a little boy is sitting in the bathtub with a yellow sponge and a little girl is (49 tokens) \\
      \end{tabular}
    }

    \caption{Duration-controlled captions generated for the same video of Spoken MiT in Fig.\ref{fig:spoken_mit_vid_cap} with ordinal embedding.
    The target duration ranges from 0.2 and 10.0 seconds.}
    \label{fig:5to100-ordinal-embed-duration-control}

\end{figure*}

\clearpage
\subsection{sampling and beam search}

As mentioned in Section \ref{sec:inference}, we used a greedy decoding strategy for caption generation since our main objective was to control caption length, rather than enhance quality metrics. Nevertheless, non-greedy decoding strategies are known to improve caption quality and here we demonstrate those decoding. Figure \ref{fig:smit_sampling_beamsearch} shows example captions generated using various decoding strategies, such as beam search (with five beams) and multinomial sampling (with $top_k=5$ and $top_p=0.9$).

The advantage of using the constant length embedding, as mentioned in Sections \ref{sec:introduction} and \ref{sec:related work}, is that it can be easily integrated into existing decoding processes. Our implementation is straightforward using the Huggingface library \cite{Wolf_EMNLP2020_Transformers_Huggingface}; the word embedding in a text decoder is replaced with an instance that returns the sum of the original word embedding and the length embedding.

Captions generated with sampling exhibit more variation and diversity compared to those generated with greedy decoding. Interestingly, this variation does not have any impact on the length of the captions, as they still align with the target length. This observation provides a strong indication that the proposed length embedding is effective in controlling the length of the caption with any decoding strategy.

\begin{figure*}

  \begin{subfigure}{\textwidth}
    \centering
    \small
    \begin{tabular}{cccp{.6\linewidth}}
      \multirow{2}{*}{decoding} &
      \multicolumn{2}{c}{length} &
      \multirow{2}{*}{captions with ordinal embedding}
      \\
       & target & gen &\\ \hline
      GT (14) &  & &
           a woman helps give her daughter a bath
           using the yellow plastic tub.  
      \\ \hline
      greedy
      & 5 & 5 &  a baby is playing. \\
      & 14 & 14 &
           a baby is playing in a small tub with a yellow rubber object.  \\
      & 20 & 20 &
           a baby is sitting in a red plastic tub with water and a pink sponge the baby is laughing. \\
       \hline
      \multirow{5}{*}{\parbox{0.1\linewidth}{beam search \\ (5 beams)}}
&5 & 5 & a baby is playing. \\
&5 & 5 & a small child playing. \\
&5 & 5 & a baby playing in. \\
&5 & 5 & a baby is taking. \\
&5 & 5 & a baby playing with. \\
&14 & 14 & this is a video of a baby playing with water in a tub. \\
&14 & 14 & this is a video of a baby that is in a little tub. \\
&14 & 14 & this is a video of a baby taking a bath in a tub. \\
&14 & 14 & this is a video of a baby that is in a red tub. \\
&14 & 14 & this is a video of a baby taking a bath in a sink. \\
&20 & 20 & this is a video of a baby that is in a red tub the baby is playing with toys. \\
&20 & 20 & this is a video of a baby that is in a red tub and the mother is washing it. \\
&20 & 20 & this is a video of a baby that is in a red tub the baby is playing with bubbles. \\
&20 & 21 & this is a video of a baby that is in a red tub and the mother is washing the baby. \\
&20 & 20 & this is a video of a baby that is in a red tub the baby is playing with water. \\        
        \hline
      \multirow{5}{*}{\parbox{0.1\linewidth}{sampling \\ ($top_k$=5)}}
&5 & 5 & mom plays with her. \\
&5 & 5 & baby in a crib. \\
&5 & 5 & a small baby laughs. \\
&5 & 6 & a video of a child. \\
&5 & 6 & a young child enjoying the. \\
&14 & 14 & a small baby can be seen as it is playing with some toys. \\
&14 & 14 & the young child playing with bubbles in the floating down in the water. \\
&14 & 14 & hey baby it's playing with this toy inside a little kitty. \\
&14 & 14 & baby sits in a bucket playing with water hose and shower head set. \\
&14 & 14 & a baby sits in a pink and yellow tub using a water pitcher. \\
&20 & 20 & the baby is sitting in the sink as a person off screen is putting the water into the tub. \\
&20 & 20 & a child with curly hair in the water with pink string floating in front of them in a tub. \\
&20 & 20 & we see several small children playing in the playroom one of the children is taking a bubble bath. \\
&20 & 20 & a woman is lying in a bassinet a baby is on her belly and playing in the water. \\
&20 & 20 & I am watching a young toddlers playing in a red bassinet the mother kisses the baby's cheek. \\
        \hline
      \multirow{5}{*}{\parbox{0.1\linewidth}{sampling \\ ($top_p$=0.9)}}
&5 & 6 & a baby is taking water. \\
&5 & 6 & a young child is taking. \\
&5 & 5 & a young toddler playing. \\
&5 & 5 & a little girl playing. \\
&5 & 5 & baby taking a bath. \\
&14 & 14 & a young baby plays with her toys while sitting in a baby tub. \\
&14 & 14 & a baby is in a little tub and is drinking from a bowl. \\
&14 & 14 & little baby is being given a bath and the sink with a sponge. \\
&14 & 14 & a child in a bassinet being given a bath by a man. \\
&14 & 14 & a baby sitting in a red bassinet is playing with water hose. \\      
&20 & 20 & two babies laying in a red plastic tub one baby is being given a bath with bubbles in it. \\
&20 & 20 & a baby is playing in a baby bathtub with bubbles as her mother's hand comes into play. \\
&20 & 20 & a baby sits in a red sink filled with water and a bottle of water she is smiling happily. \\
&20 & 20 & a baby in a playroom with a water faucet and a man holding a watering can. \\
&20 & 20 & there's a baby inside a red tube being given a bath there is a small lamp nearby also. \\
    \hline
      \end{tabular}

  \end{subfigure}

  \caption{Generated captions for the same video of Spoken MiT in Fig.\ref{fig:spoken_mit_vid_cap} with ordinal embedding. Five captions are shown for each decoding strategy.
  Results for the greedy decoding are the same as in Fig. \ref{fig:spoken_mit_vid_cap}.
  }
  \label{fig:smit_sampling_beamsearch}

\end{figure*}

\clearpage
\subsection{Comparison with LLM}

GPT-4o \cite{GPT-4o_2024_05_13} is the successor to GPT-4 \cite{openai2024gpt4technicalreport} and
is currently considered one of the state-of-the-art multimodal models. This table shows the captions generated by GPT-4o for the single video used in the experiments. After uploading the video, we used the prompt;
``\textit{describe the content of the video in N words, not the video file or format, for N=1,2,...,19,20,30,40,...,100}.'' If it failed to generate captions for some target lengths, we again generated captions for missing target lengths.
Unlike the proposed method, the sentences were generated by specifying the number of words, so the target lengths are represented here by the number of words, not the number of tokens.

Compared to the results with ordinal embedding in Figure \ref{fig:5to100-ordinal-embed-smit}, the lengths of the generated captions do not match exactly the target lengths, with an error of approximately one word, for target lengths of 20 words or less. However, the proposed method often generates captions of exactly the same length when the target length is 20 or less.

Even when the target length increases, the length of the captions generated by GPT-4o almost matches the target length. These captions often include abstract sentences with many modifiers that seem to express more than the content of the scene. In contrast, the proposed method generates captions that express only the facts that can be read from the scene, as it is trained on captioning datasets. Therefore, repetitive content tends to appear when the target length is increased.

\begin{figure*}[t]
    \centering

    {\small
    \begin{tabular}{ccp{.8\linewidth}}
    \multicolumn{2}{c}{words} & \multirow{2}{*}{captions with GPT-4o} \\
      target & gen &\\ \hline
1 & 1 & Bathing \\
2 & 2 & Baby bathing \\
3 & 3 & Mother washing baby \\
4 & 4 & Mother washing baby's hair \\
5 & 5 & Baby getting bath from mother \\
6 & 6 & Mother gently washing her baby's hair \\
7 & 6 & Mother washing baby's hair in bathtub \\
8 & 7 & Mother washing her baby's hair with care \\
9 & 7 & Mother lovingly washing baby's hair in bath \\
10 & 9 & Mother gently washing her baby's hair during bath time \\
11 & 11 & A mother gently washes her baby's hair while in the bathtub. \\
12 & 13 & A mother lovingly washes her baby's hair as they sit in the bathtub. \\
13 & 13 & A mother is gently washing her baby's hair while the baby sits calmly. \\
14 & 14 & A mother lovingly washes her baby's hair as the baby sits in a bathtub. \\
15 & 14 & A mother gently washes her baby's hair while the baby sits in a bathtub. \\
16 & 16 & A mother carefully and lovingly washes her baby's hair while the baby sits in a bathtub. \\
17 & 17 & A mother gently and lovingly washes her baby's hair while the baby calmly sits in a bathtub. \\
18 & 17 & A mother lovingly and carefully washes her baby's hair while the baby sits calmly in a bathtub. \\
19 & 18 & A mother gently washes her baby's hair while the baby sits calmly in a bathtub, enjoying the bath. \\
20 & 22 & In the video, a mother is seen lovingly washing her baby's hair while the baby sits in a bathtub, enjoying the bath. \\
30 & 34 & The video captures a tender moment where a mother is washing her baby's hair. The baby is seated in a bathtub, appearing calm and relaxed as the mother gently pours water over her hair. \\
40 & 42 & The video showcases a nurturing scene where a mother is washing her baby's hair. The baby, seated in a bathtub, looks serene and content. The mother carefully pours water over the baby's hair, ensuring the baby is comfortable and enjoying the bath. \\
50 & 49 & In this heartwarming video, a mother is seen washing her baby's hair. The baby, comfortably seated in a bathtub, appears to be enjoying the experience. The mother gently pours water over the baby's head, carefully washing the hair while ensuring the baby remains calm and content throughout the bath. \\
60 & 62 & The video shows a heartwarming moment of a mother washing her baby's hair. The baby is seated in a bathtub, appearing relaxed and content. The mother gently pours water over the baby's head, carefully washing the hair. Her actions are tender and loving, ensuring the baby is comfortable and enjoying the bath. The scene beautifully captures the bond between mother and child. \\
70 & 69 & The video captures a touching scene where a mother washes her baby's hair. The baby, seated in a bathtub, looks relaxed and at ease. The mother gently pours water over the baby's head, washing the hair with care and affection. Her tender actions ensure the baby remains comfortable and enjoys the bath. This beautiful moment highlights the loving bond between mother and child, showcasing a serene and nurturing environment. \\
80 & 79 & In this heartwarming video, a mother is seen washing her baby's hair. The baby, comfortably seated in a bathtub, appears relaxed and content. The mother gently pours water over the baby's head, washing the hair with careful and loving motions. Her tender actions ensure the baby remains at ease and enjoys the experience. The scene beautifully captures the strong bond between mother and child, highlighting a serene and nurturing moment that reflects the care and affection shared between them. \\
90 & 87 & This video captures a tender and heartwarming moment of a mother washing her baby's hair. The baby, seated comfortably in a bathtub, appears relaxed and content throughout the process. The mother gently pours water over the baby's head, carefully washing the hair with loving and tender motions. Her actions ensure the baby is at ease and enjoys the experience. The scene beautifully showcases the strong bond between mother and child, highlighting a serene and nurturing environment filled with care, affection, and mutual comfort during this intimate moment. \\
100 & 108  & This video captures a touching and serene moment between a mother and her baby during bath time. The mother is washing her baby's hair, and the baby is seated in a bathtub. The baby looks relaxed and comfortable as the mother carefully and gently pours water over the baby's head, ensuring that the baby is at ease and enjoying the process. The mother's actions are tender and loving, reflecting the bond between them. The baby appears to be enjoying the experience, with the calm environment contributing to the overall peaceful and intimate atmosphere of the video. The scene is a beautiful depiction of a nurturing and caring moment. \\
      \end{tabular}
    }

    \caption{Captions generated for the same video of Spoken MiT in Fig.\ref{fig:spoken_mit_vid_cap} with the GPT-4o \cite{GPT-4o_2024_05_13}.
    The target length ranges from 1 and 100 words, not tokens.}
    \label{fig:5to100-chatGPT-4o}

\end{figure*}


\end{document}